\documentclass{article}
\usepackage{iclr2027_conference,times}

\usepackage{amsmath,amsfonts,bm}

\def\eqref#1{equation~\ref{#1}}

\def\1{\bm{1}}

\DeclareMathAlphabet{\mathsfit}{\encodingdefault}{\sfdefault}{m}{sl}
\SetMathAlphabet{\mathsfit}{bold}{\encodingdefault}{\sfdefault}{bx}{n}

\usepackage[utf8]{inputenc}
\usepackage[T1]{fontenc}
\usepackage{multirow}

\usepackage{amsmath}
\usepackage{amsfonts}
\usepackage{amssymb}

\usepackage{graphicx}
\usepackage{booktabs}
\usepackage{threeparttable}
\usepackage{capt-of}
\usepackage{fvextra}

\DefineVerbatimEnvironment{PromptBlock}{Verbatim}{
  fontsize=\footnotesize,
  breaklines=true,
  breakanywhere=true,
  breaksymbolleft={},
  breaksymbolright={}
}

\usepackage[ruled,vlined]{algorithm2e}

\usepackage{xcolor}
\usepackage{microtype}
\usepackage{url}

\usepackage{hyperref}
\usepackage[capitalise]{cleveref}

\title{GOPAgen: Codec-Aware Agentic Long-Video Understanding with Structured Memory}

\author{%
Haozhe Chi$^{1}$, Yang Jin$^{1}$, Yadong Mu$^{1}$\\
$^{1}$Peking University
}
\iclrfinalcopy

\begin{document}

\maketitle
\lhead{}

\begin{abstract}
Agentic long-video question answering often relies on sparse RGB sampling and caption retrieval. This reliance can lead to missed brief motion events and repeated processing of irrelevant intervals. We introduce GOPAgen, a framework that integrates codec-native Groups of Pictures (GOPs) and motion vectors into an agentic retrieval pipeline. GOPAgen constructs global textual memory and uses two-stage query-conditioned temporal selection. A global-sufficiency check precedes local evidence extraction. Caption and motion agents represent selected intervals as typed memory pages containing complementary appearance and motion evidence, temporal metadata, and provenance. A GOP-Tree organizes these pages for conditional retrieval when the global evidence is insufficient. GOPAgen achieves 65.3\% accuracy on MotionBench Test and 78.7\% on EgoSchema, exceeding the reported agentic baselines listed for these benchmarks. It also achieves 73.2\% on LongVideoBench validation and 77.3\% on MLVU, while remaining below the strongest listed baseline on LVBench and Video-MME Long. These results support codec-aware structured memory as an effective interface for selective long-video reasoning.
\end{abstract}

\section{Introduction}
Long-form video question answering requires a model to identify evidence distributed across extended temporal contexts while preserving the local details needed to answer a query. This setting is challenging because visual events, narratives, and object interactions evolve at different time scales, whereas the number of frames grows linearly with video duration. Although recent multimodal large language models (MLLMs) support increasingly long contexts~\citep{cheng2025vilamp,song2024moviechat,damonlpsg2025videollama3,liu2025videomind,liu2024oryx}, supplying a densely decoded long video remains costly and can introduce substantial redundancy. Consequently, long-video systems commonly sample frames, compress visual tokens, or search for query-relevant temporal regions~\citep{song2024moviechat,li2024llava,wu2024longvideobench,fu2025video,chen2024timemarker,chen2024longvila}.

Agentic methods make this selection process explicit. They use language or multimodal models to iteratively localize, inspect, and aggregate evidence from a long video~\citep{luo2024videorag,zuo2025videolucy,zhao2024longagent,zhang2025deep,wang2025videotree}. However, many such pipelines represent local evidence primarily through isolated RGB frames and frame-level captions. Sparse sampling can therefore miss a short-lived event, while caption histories provide only an indirect account of displacement and temporal change. Compressed video offers a complementary source of temporal evidence. Groups of Pictures (GOPs) associate intra-coded reference frames with predictively coded frames. Their block motion vectors describe the encoder's temporal predictions relative to reference frames. Motion vectors are not optical flow and do not by themselves provide semantic understanding, but they offer codec-native direction and magnitude cues that complement RGB appearance within a selected interval.

A second challenge is to keep this heterogeneous evidence aligned during retrieval and reasoning. Existing long-video agents already use coarse-to-fine search, captioning, vector retrieval, memory, and language-model reasoning~\citep{wang2025videotree,yang2025vca,zuo2025videolucy,zhang2025deep}. Our contribution is therefore not that these individual operations are new. The systems question is how to connect codec-level motion evidence to agentic reasoning while preserving its temporal origin. The interface must also avoid exposing raw motion-vector arrays directly to the final language model. A suitable interface must preserve the selected interval, its GOP provenance, the RGB appearance description, the motion description, and the representation used for retrieval. It must also distinguish a global summary of the complete video from query-selected local evidence.

\begin{figure}[t!]
    \centering
    \includegraphics[width=0.98\linewidth]{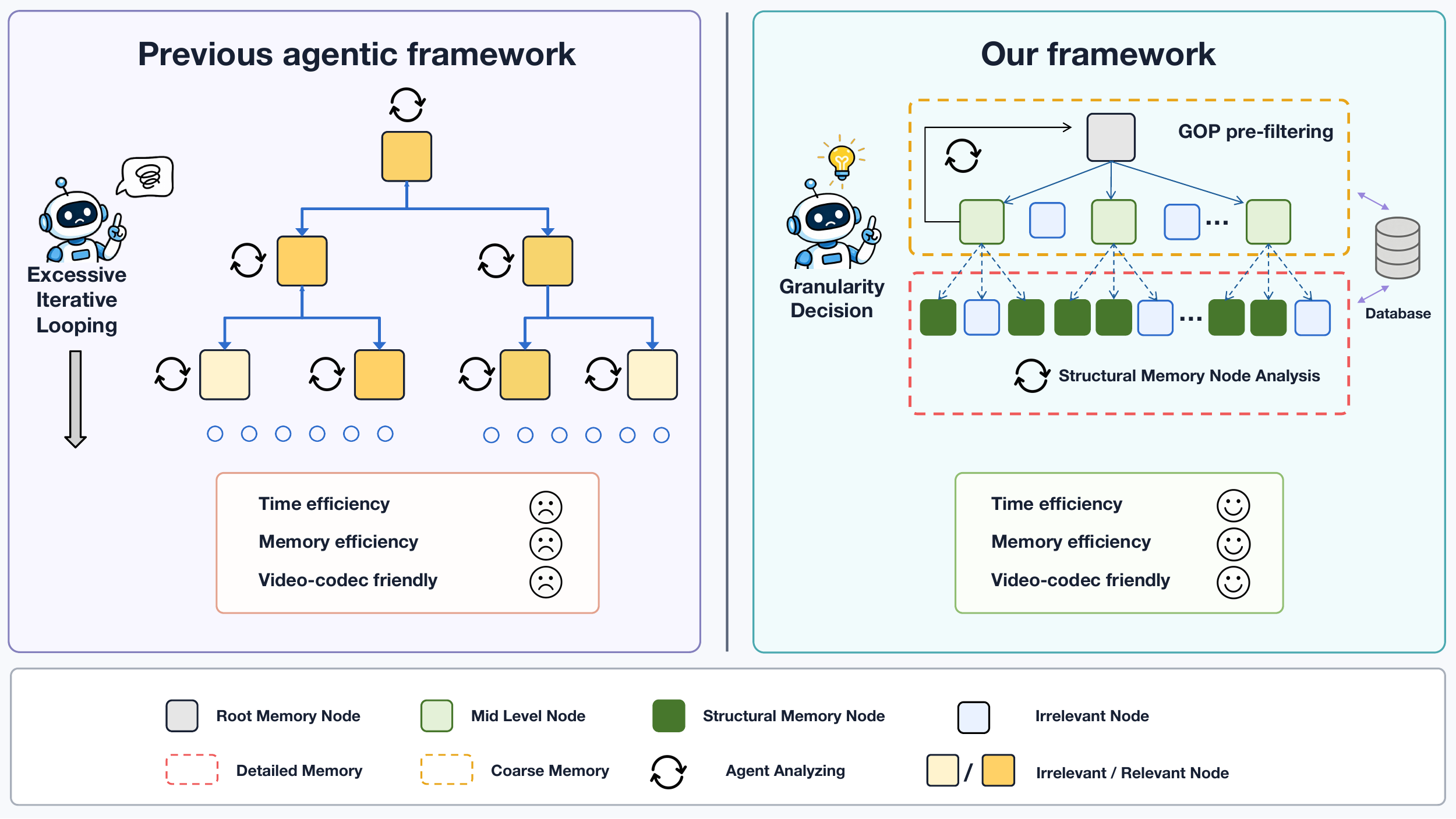}
    \vspace{-0.1in}
    \caption{Conceptual comparison of iterative sparse-RGB search and GOPAgen. \textbf{Left:} An agent iteratively selects RGB observations and appends captions to its interaction history. Fragmented evidence can require repeated inspection and increase retained context. \textbf{Right:} GOPAgen selects coarse and fine intervals using sparse global observations and the query. For queries that fail the global-sufficiency check, caption and motion agents produce interval-aligned descriptions stored with temporal metadata and provenance in typed memory pages. The GOP-Tree organizes these pages for conditional retrieval.}
    \label{fig:architectureorig}
    \vspace{-0.2in}
\end{figure}

We address this question with GOPAgen, a system-level framework for codec-aware agentic long-video understanding. As illustrated in~\cref{fig:architectureorig}, GOPAgen first constructs a global textual memory from sparsely sampled RGB frames and performs two query-conditioned temporal-selection rounds: coarse interval selection followed by fine interval refinement. Before local processing, the reasoning agent checks whether the global memory already supports an answer. Only queries requiring additional evidence trigger local-agent calls on the retained fine intervals. A caption agent produces an appearance description, whereas a dedicated motion agent jointly interprets the associated GOP reference frames and motion vectors to produce a textual motion description. GOPAgen stores both descriptions, the temporal interval, GOP identifiers, a retrieval embedding, and provenance in a typed memory page. The resulting pages are organized in an interval hierarchy, termed the GOP-Tree. For this local-evidence path, descriptions and pages are generated on demand after the query arrives. The system retrieves, ranks, and compresses the resulting pages before producing the answer.

Together, these components connect codec-native motion evidence with structured agent memory. The motion agent converts raw motion vectors into textual descriptions, which are organized by the page index and consumed by the reasoning agent.

Our contributions are summarized as follows:
\begin{itemize}
    \item We introduce a codec-aware interface and train a motion agent to jointly interpret GOP reference frames and motion vectors, producing interval-aligned textual motion evidence.
\item We organize appearance and motion evidence, timestamps, GOP identifiers, embeddings, and provenance into typed memory pages and an interval-based GOP-Tree for conditional retrieval.
\item We evaluate the complete system and standalone motion agent across video-QA benchmarks, with supplementary component comparisons, retrieval diagnostics, and resource measurements.
\end{itemize}

\section{Related Work}
\paragraph{Multimodal Models for Video Understanding.}
Video MLLMs extend image--language models to temporal inputs~\citep{liu2023visual,ye2024mplug,shao2025growing,yu2025prophet,shao2025imp,openai2025o3,achiam2023gpt,openai2025gpt4.1,bai2025qwen2,damonlpsg2025videollama3,wang2025internvideo2}. Context extension~\citep{team2023gemini,wei2025visual,zhang2024long}, visual-token compression~\citep{yang2025pvc,tao2025dycoke,shen2024longvu}, and audiovisual integration~\citep{cappellazzo2025large,shu2025audio,xu2025qwen2} provide complementary approaches to processing long videos. Frame-based encoders convert selected observations into tokens for language-model reasoning~\citep{xu2025qwen2,wang2024qwen2,li2024llava,li2024llavaov}. Sampling methods allocate observations to informative intervals~\citep{song2025moviechat+,yao2025generative,tang2025adaptive}, while token compression and memory mechanisms reduce the retained representation~\citep{song2024moviechat,bolya2022token,li2024videochat,wang2025adaretake,kahatapitiya2025language}. These choices address different stages of the visual-input pipeline. Sampling determines which observations enter the model, whereas compression determines how much information is retained from those observations. Memory provides a further interface for carrying selected information into later reasoning steps. Their temporal coverage and retained detail therefore depend on both the sampling policy and the representation stored for subsequent use. GOPAgen complements these approaches by incorporating codec motion evidence alongside RGB descriptions in interval-aligned memory pages.

\paragraph{Agentic Video Understanding.}
Agentic systems coordinate evidence acquisition and reasoning through explicit search and tool use~\citep{zuo2025videolucy,fan2024videoagent,luo2024videorag,yao2023react}. Examples include VideoAgent~\citep{wang2024videoagent}, DrVideo~\citep{ma2024drvideo}, and VideoTree~\citep{wang2025videotree}. These systems differ in their search policies, temporal granularity, and representation of retrieved observations~\citep{chen2025lvagent,li2024searchlvlms,zhao2024longagent,he2024ma,yang2025vca}. Clip-level retrieval and iterative exploration support localized inspection~\citep{zhang2025deep,pang2025mr,jeong2025videorag}, while VideoARM combines adaptive reasoning with dynamic memory~\citep{yin2025videoarm}. Search policy and evidence representation are related but distinct design choices. A system must determine both where to inspect and how to preserve the resulting observations for later reasoning. GOPAgen retains this separation by using two-level temporal selection to identify intervals and typed pages to store their appearance and motion descriptions. Rather than introducing search or memory in isolation, it studies their integration with codec-aware motion descriptions. Typed pages preserve the evidence type and temporal provenance, and the GOP-Tree provides a fixed-depth organization for conditional retrieval.

\section{Method}

\subsection{Problem Formulation and Overview}
Given a video $V$ and a natural-language question $q$, the goal is to predict an answer from the visual evidence relevant to the question. GOPAgen accesses two complementary representations of $V$. The decoded RGB stream preserves objects, scenes, and appearance changes, but densely processing it is costly for long videos. The compressed bitstream organizes the same video into an ordered sequence of Groups of Pictures (GOPs) and exposes codec-native motion information. We denote the $j$-th GOP by $B_j$. Each GOP record stores its temporal extent, an intra-coded reference frame $K_j$, the associated motion-vector records $\mathrm{MV}_j$, and the codec metadata required to interpret them. Individual vector records contain a block location, a coded displacement, reference-direction information, and a timestamp. These block-level fields are used during motion tokenization but are not manipulated separately by later stages.

The two representations provide different evidence. RGB frames describe what appears in an interval, whereas codec motion vectors describe how the encoder predicts blocks across frames. Because these vectors depend on the codec and encoding configuration, they are not equivalent to optical flow and are not interpreted in isolation. GOPAgen instead combines them with the corresponding GOP reference frames and translates the paired evidence into text that can be retrieved by the downstream reasoning pipeline.

\begin{figure*}[t]
    \centering
    \includegraphics[width=0.95\linewidth]{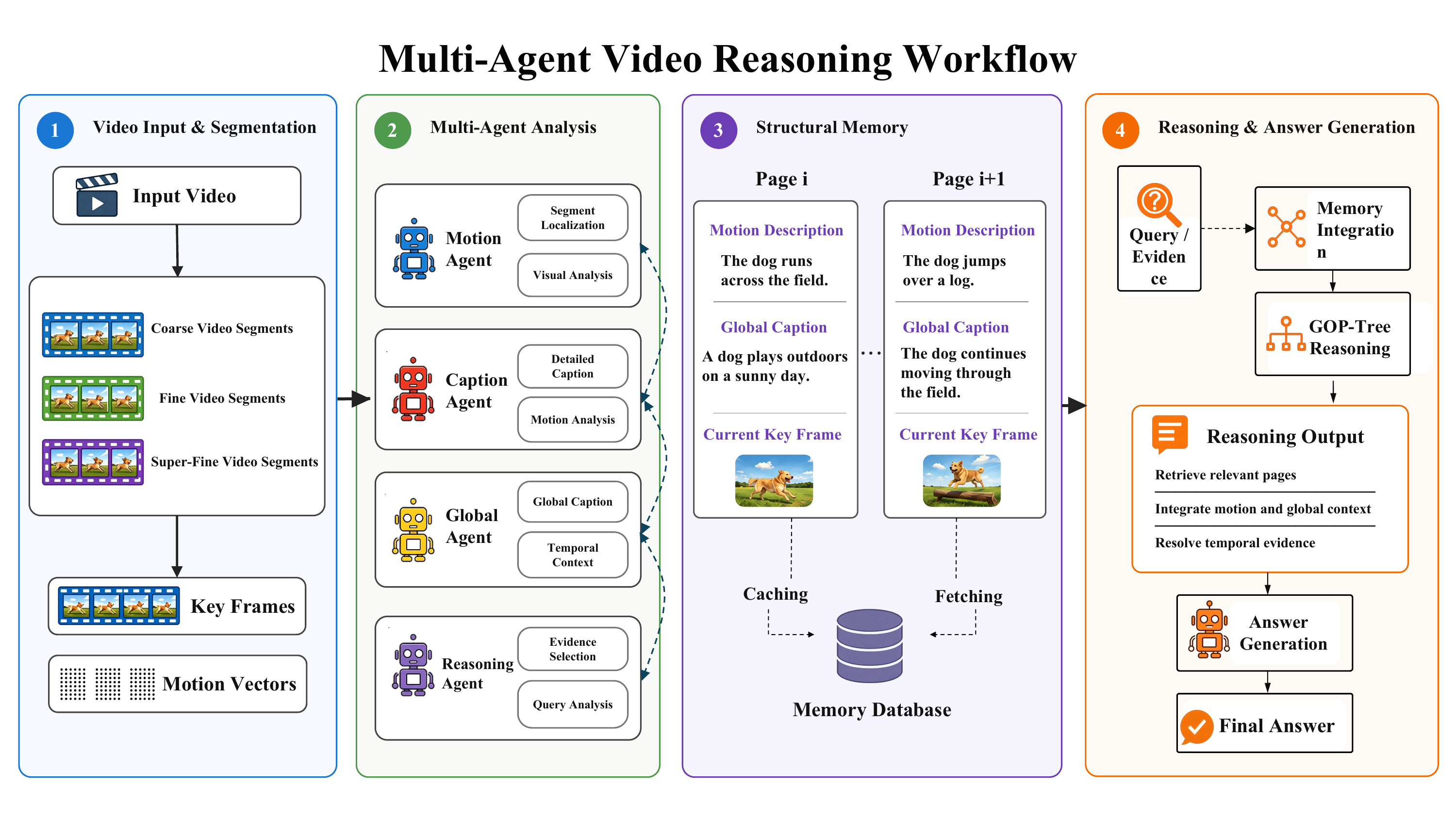}
    \caption{Functional components of GOPAgen, arranged from left to right. \textbf{(1) Video input and segmentation:} temporal regions, reference frames, and motion-vector records provide the source evidence. \textbf{(2) Multi-agent analysis:} global, caption, motion, and reasoning agents perform complementary tasks. \textbf{(3) Structured memory:} pages associate textual descriptions with visual references and support caching and retrieval. \textbf{(4) Reasoning and answer generation:} retrieved evidence is integrated through the GOP-Tree to produce an answer. The global-sufficiency gate precedes local-agent calls and page construction, as specified in~\cref{alg:gop_tree_reasoning}.}
    \label{fig:illustration}
\end{figure*}

As illustrated in~\cref{fig:illustration}, the inference procedure follows a coarse-to-fine evidence flow. First, the global agent builds a time-aware textual memory of the video and uses it to retain query-relevant coarse and fine intervals. Before local processing, the reasoning agent checks whether global memory already supports an answer. If global evidence is insufficient, the caption and motion agents extract complementary descriptions from the retained fine intervals. These descriptions and their temporal and codec provenance are then stored as typed memory pages. Finally, the GOP-Tree connects the global memory to these newly constructed pages for retrieval and final reasoning. Queries answered from global memory skip local description generation, page construction, and retrieval. A page index supports storage and retrieval but does not perform reasoning. The following subsections describe these stages in inference order. Model configurations, codec settings, and retrieval budgets appear in~\cref{app:implementation_details}, with agent prompts in~\cref{app:implementation_prompts}.

\subsection{Query-Conditioned Two-Level Temporal Selection}
The first stage determines where detailed local processing is needed. GOPAgen sparsely samples a bounded set of timestamped RGB frames from $V$. Given these frames and $q$, the global agent produces a coarse, time-aware textual memory $\mathcal{G}$. This memory provides a low-resolution account of the complete video and supplies the context used to evaluate candidate intervals. For questions whose evidence is already represented by the sampled frames, $\mathcal{G}$ may also be sufficient for final reasoning without local retrieval.

Temporal selection proceeds in two rounds with a fixed search depth. The first round divides the video timeline into coarse candidates and evaluates their relevance to $q$ in the context of $\mathcal{G}$. The retained coarse intervals form $\mathcal{U}_{c}$. Each interval in $\mathcal{U}_{c}$ is then partitioned into shorter candidates, which are evaluated in a second round. The retained fine intervals form $\mathcal{U}_{f}$. We use notation only for the two retained sets because candidate construction and intermediate scores are not consumed by later modules.

The two rounds balance temporal coverage and local precision. Coarse selection limits detailed inspection to a small portion of the video, while fine selection identifies intervals at the granularity required by the local agents. Non-contiguous intervals remain separate because evidence for one question may occur at multiple distant times. When the global-sufficiency check requests additional evidence, only intervals in $\mathcal{U}_{f}$ proceed to local evidence extraction. Thus, the selection stage converts the whole video into a bounded set of query-dependent inputs for the caption and motion agents described next. The global sampling rule, interval durations, subdivision procedure, scoring method, retention budgets, overlap handling, and token limits are specified in~\cref{app:selection_details}.

\subsection{Codec-Aware Motion Agent}
Once the fine intervals have been selected, GOPAgen augments their RGB appearance with codec-level motion evidence. This step addresses events whose temporal change may not be adequately represented by the sparse frames used to construct $\mathcal{G}$. For a retained interval $u\in\mathcal{U}_{f}$, let $\mathcal{B}_u$ denote the indices of GOPs whose temporal ranges overlap $u$. The associated motion records are restricted to the timestamps of $u$, while the reference frames supply their appearance context. With this restriction implicit in the inputs, the motion agent produces
\begin{equation}
m_u=\operatorname{MotionAgent}\!\left(\{(K_j,\mathrm{MV}_j):j\in\mathcal{B}_u\}\right),
\label{eq:motion_agent}
\end{equation}
where $m_u$ is a textual description of motion within the selected interval. The reference frames supply the appearance context needed to interpret the vectors. Internally, the visual encoder processes the reference frames, while the motion tokenizer normalizes and tokenizes the corresponding vector records. The motion agent jointly interprets both streams under a motion-description instruction. Its output is a fine-grained motion description, not a separate answer to $q$. Internal features, motion tokens, and prompt variables are transient. Only $m_u$ is passed to the memory-construction stage.

The motion agent follows the LLaVA-NeXT-Video pipeline~\citep{li2024llavaov}, using Qwen2~\citep{xu2025qwen2} as the language backbone and SigLIP-2~\citep{tschannen2025siglip} as the vision encoder. Training proceeds through image--text pretraining, mixed image/video instruction training, and motion-focused fine-tuning. The final stage introduces paired GOP reference frames and motion vectors and includes additional motion-caption examples from LLaVA-Hound. Complete data mixtures, optimization schedules, hardware, and frozen or trainable modules are listed in~\cref{app:motion_training_details}.

Motion vectors are extracted once per video and assigned stable video and GOP identifiers. At inference time, the timestamps of $u$ are used to locate the relevant records without repeating motion-vector extraction for every question. The implementation accounts for frame types, prediction-reference direction, coordinate normalization, vector tokenization, and intervals spanning multiple GOPs. These codec-dependent rules are specified in~\cref{app:codec_details}. The resulting description $m_u$ provides the motion field of the typed page constructed in the next stage.

\subsection{Typed Memory Pages}
The temporal-selection stage identifies where to inspect, and the local agents describe what appears and how it changes. GOPAgen next binds these outputs to their temporal and codec origins so that retrieval does not discard their context. For a query requiring local evidence, the caption agent describes the RGB observations in each selected interval $u$ and produces an appearance description $c_u$. The motion agent supplies the corresponding description $m_u$ from~\cref{eq:motion_agent}. The two fields remain separate: $c_u$ records objects, scenes, and visual states, whereas $m_u$ records codec-supported temporal change.

We represent the resulting typed page as
\begin{equation}
P_u=(u,\mathcal{B}_u,c_u,m_u,z_u,\pi_u),
\label{eq:memory_page}
\end{equation}
where $z_u$ is a retrieval embedding computed from the appearance and motion text, and $\pi_u$ is a provenance record. The provenance record stores the video identifier, interval timestamps, parent interval, and the models and prompts used to generate the page. The pages constructed for $q$ form the query-conditioned local memory $\mathcal{M}_q$. This representation exposes the evidence type, temporal scope, codec source, and generation history to the retrieval pipeline without placing intermediate model features in the reasoning context.

The page index and the underlying media stores serve different purposes. The index stores $z_u$ together with the page and interval metadata. Retrieval first filters entries by video identity and temporal constraints and then ranks the eligible pages using their stored representations. Decoded RGB observations and raw motion-vector arrays remain in their source stores, while $\mathcal{B}_u$ links a page to the required codec records. Inserting or updating a page therefore changes the textual memory and its metadata without rewriting the source vectors. The embedding model, page schema, persistence, interval filtering, caching, duplicate handling, and update semantics are detailed in~\cref{app:index_details}. The resulting collection $\mathcal{M}_q$ provides the local evidence attached to the GOP-Tree.

\subsection{GOP-Tree and Conditional Reasoning}
For queries requiring local evidence, the final stage connects global memory $\mathcal{G}$ with the pages generated on demand in $\mathcal{M}_q$. The GOP-Tree is a fixed interval hierarchy rather than a learned graph. Its root represents the complete video and stores $\mathcal{G}$. The next level contains the retained coarse intervals $\mathcal{U}_{c}$, and the leaves contain the retained fine intervals $\mathcal{U}_{f}$ and reference their corresponding pages. Parent--child links indicate temporal containment. Non-contiguous selections remain separate siblings, and GOP records are referenced through $\mathcal{B}_u$ rather than represented as additional tree nodes.

The global-sufficiency check precedes caption-agent and motion-agent calls, page construction, and local retrieval. The reasoning agent first determines whether $\mathcal{G}$ provides sufficient evidence for $q$. If it does, the agent answers from global memory without constructing local pages. Otherwise, GOPAgen generates local descriptions, writes their pages, and constructs the GOP-Tree for the query. The local ranker then selects relevant branches and pages from this tree. The retained page text is then compressed and combined with $\mathcal{G}$ to form the final reasoning context. Motion affects the answer through $m_u$ and the page embedding $z_u$. The reasoning agent never receives raw motion-vector arrays. Full traversal pseudocode and the sufficiency, branching, ranking, compression, duplicate-handling, and decoding settings are reported in~\cref{app:reasoning_details,alg:gop_tree_reasoning}.

This conditional procedure links the output of each preceding stage: $\mathcal{G}$ supplies whole-video context, $\mathcal{U}_{c}$ and $\mathcal{U}_{f}$ define the hierarchy, and $\mathcal{M}_q$ supplies interval-specific appearance and motion evidence. The two-level policy bounds search depth and restricts local-agent processing to selected intervals. The global-only path still incurs global inspection, sufficiency assessment, and any video preprocessing already performed. Consequently, bounding local inspection alone does not establish lower end-to-end cost. Our resource analysis therefore accounts for decoding, motion-vector extraction, model calls, page and index construction, retrieval, compression, API usage, and storage (\cref{app:end_to_end_cost}).

\section{Experiments}
\subsection{Experimental Setup}
GOPAgen consists of four model agents: a global agent, a caption agent, a motion agent, and a reasoning agent. A local ranker and page index support evidence storage and retrieval. Qwen2.5-VL~\citep{Bai2025Qwen25VLTR} performs global inspection and temporal selection, AuroraCap~\citep{chai2024auroracap} supplies appearance descriptions, and our trained motion agent supplies motion descriptions. DeepSeek-V3~\citep{liu2024deepseek} performs final reasoning through its API.

The motion agent follows the LLaVA-NeXT-Video training pipeline~\citep{li2024llavaov}, using Qwen2~\citep{xu2025qwen2} and SigLIP-2~\citep{tschannen2025siglip} as its language and vision backbones. Motion-focused fine-tuning uses 20,000 LLaVA-Hound motion-caption examples generated using GPT-4o~\citep{achiam2023gpt}. Training data and optimization settings are specified in~\cref{app:motion_training_details}.

ChromaDB~\citep{vijay2025automating} stores retrieval embeddings and page metadata. RGB observations and motion-vector arrays remain in their source stores and are resolved through video identifiers, GOP identifiers, and timestamps. During conditional retrieval, interval metadata restricts the eligible pages to the relevant video and temporal regions. Updating a textual page does not require rewriting the source motion-vector arrays. Index and storage settings appear in~\cref{app:index_details}.

\subsection{Evaluation Benchmarks and Metrics}
We evaluate the complete system on Video-MME~\citep{fu2025video}, MotionBench~\citep{hong2025motionbench}, NExT-QA~\citep{xiao2021next}, LongVideoBench~\citep{wu2024longvideobench}, LVBench~\citep{wang2025lvbench}, EgoSchema~\citep{mangalam2023egoschema}, and MLVU~\citep{zhou2025mlvu}. Together, these benchmarks cover motion understanding, temporal reasoning, and long-video question answering. Standalone motion-agent evaluation additionally includes MSVD-QA~\citep{chen2011collecting}, MSRVTT-QA~\citep{xu2016msr}, and ActivityNet-QA~\citep{yu2019activitynet}. In our evaluations, questions are presented in multiple-choice format, and accuracy is computed by matching the selected option to the ground-truth option. Tables report accuracy (\%). Benchmark splits and Video-MME subtitle settings are indicated in the corresponding headers.
\subsection{Quantitative Results}
\subsubsection{Agentic Framework Results}
Tables~\ref{tab:111} and~\ref{tab:222} compare GOPAgen with the listed video-understanding systems. On Video-MME, GOPAgen achieves 84.5\% and 75.1\% accuracy on the short and medium subsets, respectively, and 75.4\% overall. Its MotionBench Test accuracy is 65.3\%, compared with 63.4\% for Deep Video Discovery, while its EgoSchema accuracy is 78.7\%, compared with 76.6\% for Deep Video Discovery. These are the highest reported values among the listed methods for these columns.

On LongVideoBench validation, GOPAgen achieves 73.2\% overall and 69.7\% on the long subset, compared with 71.6\% and 68.6\% for Deep Video Discovery. GOPAgen also achieves 77.3\% on MLVU, compared with 76.1\% for VideoLucy. The advantage is not uniform: GOPAgen trails Deep Video Discovery on LVBench (73.5\% versus 74.2\%) and Video-MME Long (66.7\% versus 67.3\%). These comparisons describe system-level performance under the reported settings. Component comparisons are presented separately in~\cref{app:component_attribution}.
\subsubsection{Standalone Motion-Agent Results}
Table~\ref{tab:333} evaluates the motion agent independently of the complete retrieval pipeline. It achieves the highest reported accuracy among the listed methods on MSVD-QA, MSRVTT-QA, MotionBench, and ActivityNet-QA. Relative to ReMoRA, its accuracy is higher on MSVD-QA (77.5\% versus 73.1\%) and ActivityNet-QA (61.7\% versus 60.5\%). ReMoRA performs better on NExT-QA (84.2\% versus 83.1\%).

\begin{table*}[t]
    \centering
    \caption{Video-question-answering accuracy (\%) on Video-MME, MotionBench, and EgoSchema. Video-MME results exclude subtitles. Short, Medium, and Long denote its duration subsets. MotionBench reports development and test results. Bold marks the highest value in each column among the listed methods. }
    \label{tab:111}
    \renewcommand{\arraystretch}{0.85}
    \setlength{\tabcolsep}{3pt}
    \resizebox{0.99\linewidth}{!}{
    \begin{threeparttable}
        \begin{tabular}{l ccccccc}
            \toprule
            {\textbf{Methods}} &  \multicolumn{4}{c}{\textbf{Video-MME (w/o sub)}}    & \multicolumn{2}{c}{\textbf{MotionBench}} & {\textbf{EgoSchema}} \\
            \cmidrule(lr){2-5} \cmidrule(lr){6-7}
            ~   & \textit{Short} & \textit{Medium} & \textit{Long}  & \textit{Overall} &  \textit{Dev} & \textit{Test} & ~ \\
            \midrule
            \multicolumn{8}{l}{\textcolor{gray}{\textit{Closed-Source Models}}} \\
            Gemini-2.0-Flash~\citep{team2023gemini}    & - & - & 63.0 & - & - & - & 71.2 \\
            Gemini-1.5-Pro~\citep{team2023gemini}  & 81.7 & 74.3 & \textbf{67.4} & 74.5 & 51.0 & 50.0 & 71.1\\
            OpenAI o3~\citep{openai2025o3} & - & - & 63.2 & - & - & - & 63.2\\
            GPT-4o~\citep{achiam2023gpt}  & 80.0 & 70.3 & 65.3 & 71.9 & - & - & 72.2\\
            \midrule
            \multicolumn{8}{l}{\textcolor{gray}{\textit{Open-Source Models}}} \\
            Qwen2.5-VL-72B~\citep{bai2025qwen2}    & - & - & - & 73.3 & 56.14 & 56.1 & 76.2\\
            mPLUG-Owl3~\citep{ye2024mplug} &  70.0 & 57.7 & 50.1 & 59.3 & - &  - & - \\
            InternVL2.5-78B~\citep{wang2025internvideo2}   & 82.8 & 70.9 & 62.6 & 72.1 & 60.9 & 61.0 & -\\
            AdaReTaKe~\citep{wang2025adaretake}  & 80.6 & 74.9 & 65.0 & 73.5 & - &  - & -\\
            \midrule
            \multicolumn{8}{l}{\textcolor{gray}{\textit{Agentic Video-Understanding Frameworks}}} \\
            VideoAgent~\citep{wang2024videoagent}  & - & - & - & - & -  &  - & 60.2\\
            VideoAgent~\citep{fan2024videoagent}  & - & - & - & - & - & - & 63.2 \\
            VideoTree~\citep{wang2025videotree}  & - & - & 54.2 & - & 52.9  &  53.1 & 54.2\\
            VCA~\citep{yang2025vca}   & - & - & - & - & 52.5  & 51.3 & 73.6 \\
            Logic-in-Frames~\citep{guo2025logic}  & 71.9 & 61.9 & 55.2 & 63.0 & - & - & - \\
            MR. Video~\citep{pang2025mr}   & - & - & 61.8 & - & - & - & 73.0\\
            VideoLucy~\citep{zuo2025videolucy}   & 78.6 & 72.1 & 66.8 & 72.5 & 58.7 & 59.1 & -\\
            Deep Video Discovery~\citep{zhang2025deep}   & - & - & 67.3 & - & 62.7 & 63.4 & 76.6\\
            \textbf{GOPAgen (Ours)} & \textbf{84.5} & \textbf{75.1} & 66.7 & \textbf{75.4} & \textbf{64.5} & \textbf{65.3} & \textbf{78.7} \\
            \bottomrule
        \end{tabular}
    \end{threeparttable}
    }
    \label{tab:video_benchmarks}
\end{table*}

\begin{table*}[t]
    \caption{Accuracy (\%) on long-video benchmarks. LongVideoBench reports validation-set performance overall and on its long-video subset. Video-MME reports the long subset without subtitles. LVBench and MLVU report overall scores. Bold marks the highest value among the listed methods for each column.}
    \label{tab:222}

    \centering
    \renewcommand\arraystretch{0.85}
    \setlength{\tabcolsep}{3pt}
    \resizebox{0.99\linewidth}{!}{
    \begin{threeparttable}
        \begin{tabular}{l ccccc}
            \toprule
            \multirow{2}{*}{\textbf{Methods}}   & \textbf{LVBench} & \multicolumn{2}{c}{\textbf{LongVideoBench (Val)}}   & \textbf{Video-MME}    & \textbf{MLVU} \\
            ~       & \textbf{Overall} & \textbf{Overall} & \textbf{Long}  & \textbf{Long (w/o sub)}  & \textbf{Overall} \\
            \midrule
            \multicolumn{6}{l}{\textcolor{gray}{\textit{Closed-Source Models}}} \\
            Gemini-1.5-Pro~\citep{team2023gemini}  & 33.1 & 64.0 & 58.6 & \textbf{67.4}  & -  \\
            Gemini-2.0-Flash~\citep{team2023gemini}  & 48.3 & - & 45.7 & 63.0  & -  \\
            GPT-4o~\citep{achiam2023gpt} & 48.9  & 66.7 & 60.9 & 65.3  & 64.6  \\
            OpenAI o3~\citep{openai2025o3} & 57.1  & 67.5 & 60.6 & 63.2 & - \\
            \midrule
            \multicolumn{6}{l}{\textcolor{gray}{\textit{Open-Source Models}}} \\

            mPLUG-Owl3~\citep{ye2024mplug} & 43.5  & 59.8 & - & 50.1  & -  \\
            InternVL2.5-78B~\citep{wang2025internvideo2} & 43.6  & 63.6 & - & 62.6 & 75.7 \\
            Qwen2.5-VL-72B~\citep{bai2025qwen2} & 47.7  & 60.7 & - & -  & 74.6  \\
            AdaReTaKe~\citep{wang2025adaretake} & 53.3  & 67.0 & - & 65.0 & -  \\
            \midrule
            \multicolumn{6}{l}{\textcolor{gray}{\textit{Agentic Video-Understanding Frameworks}}} \\
            VideoTree~\citep{wang2025videotree} & 28.8 &  - & - & 54.2   & 60.4 \\
            VideoAgent~\citep{wang2024videoagent} & 29.3  & - & - & - & -  \\
            VideoRAG~\citep{luo2024videorag} & - & - & - & - & 72.4 \\
            VCA~\citep{yang2025vca} & 41.3 & - & - & -  & -  \\
            MR. Video~\citep{pang2025mr} & 60.8 & - & 61.6 & 61.8 & -  \\
            VideoLucy~\citep{zuo2025videolucy} & 58.8 & 70.2 & - & 66.8 & 76.1 \\
            Deep Video Discovery~\citep{zhang2025deep} & \textbf{74.2} &  71.6 & 68.6  & 67.3 & -  \\
            \textbf{GOPAgen (Ours)} & 73.5 & \textbf{73.2} & \textbf{69.7} & 66.7 & \textbf{77.3} \\
            \bottomrule
        \end{tabular}
    \end{threeparttable}
    }
    \label{tab:long_video_benchmarks}

\end{table*}

\begin{table*}[t]
\centering
\caption{Standalone motion-agent accuracy (\%) compared with video VLMs and codec-based video models. MSVD-QA, MSRVTT-QA, and ActivityNet-QA evaluate video question answering. MotionBench entries report development/test accuracy. These results exclude the full GOPAgen retrieval pipeline. Bold marks the highest value in each column among the listed methods, including both MotionBench splits. }
\label{tab:333}
\renewcommand{\arraystretch}{1.1}
\resizebox{\linewidth}{!}{
\begin{tabular}{lccccc}
\toprule
\multirow{2}{*}{\textbf{Methods}} & \textbf{MSVD-QA} & \textbf{MSRVTT-QA} & \textbf{NExT-QA} & \textbf{MotionBench} & \textbf{ActivityNet-QA} \\
~ & ~ & ~ & ~ & (Dev/Test) & ~ \\
\midrule
\multicolumn{6}{l}{\textcolor{gray}{\textit{Open-Source Video VLMs}}} \\
LLaVA-NeXT-Video~\citep{li2024llava} & 71.3 & 56.5 & 77.3 & 48/40 & 47.6 \\
Video-LLaVA (EMNLP 2024)~\citep{lin2024video} & 70.7 & 59.2 & -- & -- & 45.3 \\
LLaMA-VID (ECCV 2024)~\citep{li2024llama} & 69.7 & 57.7 & -- & -- & 47.4 \\
MovieChat (CVPR 2024)~\citep{song2024moviechat} & 75.2 & 52.7 & -- & -- & 45.7 \\
VideoChat2 (CVPR 2024)~\citep{li2024mvbench} & 70.0 & 54.1 & 61.7 & -- & 49.1 \\
PLLaVA~\citep{xu2024pllava} & 76.6 & 62.0 & -- & 52/51 & 56.3 \\
Video-LLAMA2~\citep{cheng2024videollama} & 70.5 & -- & -- & -- & 50.3 \\
\midrule
\multicolumn{6}{l}{\textcolor{gray}{\textit{Codec-Based Video VLMs}}} \\
VideoLaVIT (ICML 2024)~\citep{jin2024video} & 73.2 & 59.3 & -- & 40.1/40.8 & 50.1 \\
EMA (CVPR 2025)~\citep{zhao2025efficient} & 75.8 & 58.5 & -- & -- & 52.1 \\
ReMoRA (CVPR 2026)~\citep{yashima2026remora} & 73.1 & -- & \textbf{84.2} & -- & 60.5 \\
\midrule
\textbf{Motion Agent (Ours)} & \textbf{77.5} & \textbf{65.1} & 83.1 & \textbf{55.7/55.5} & \textbf{61.7} \\
\bottomrule
\end{tabular}
}
\label{tab:video_qa}
\end{table*}

\subsection{Qualitative Comparison}
Figure~\ref{fig:333} compares GOPAgen and VideoLucy responses on questions about event counts, actions, and relative motion. The panels show answer differences rather than a step-by-step retrieval trace. Figures~\ref{fig:app1}--\ref{fig:app4} provide additional answer comparisons with VideoLucy and Deep Video Discovery. Figure~\ref{fig:444} illustrates global-to-local memory construction for a tutorial video. These examples illustrate individual reasoning outcomes but do not establish how frequently such outcomes occur across the evaluation set.
\subsection{Ablation Analysis}
\begin{figure}[t]
    \centering
    \begin{minipage}{0.48\linewidth}
        \centering
        \includegraphics[width=\linewidth]{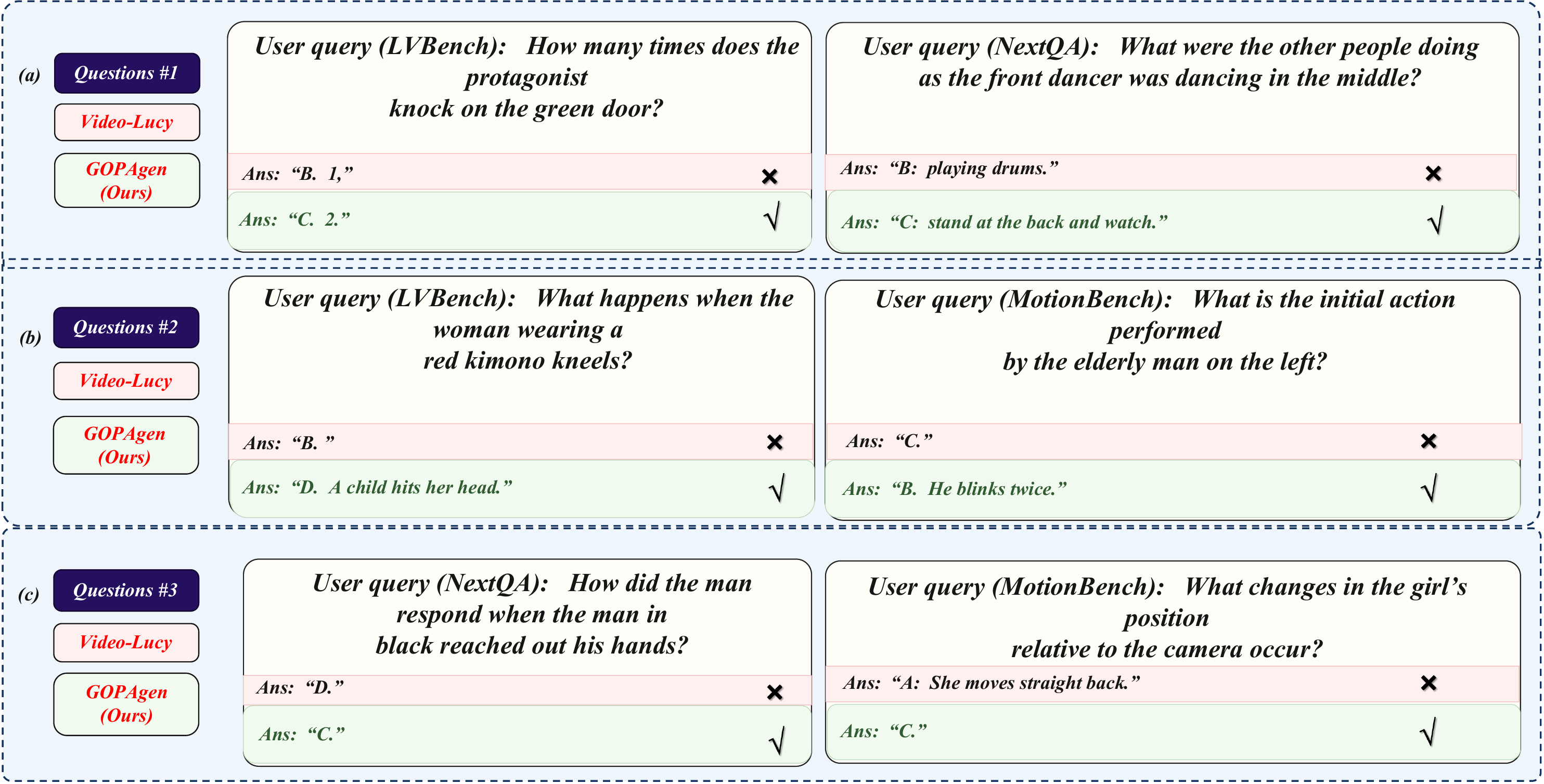}
        \vspace{-0.1in}
        \caption{Qualitative answer comparisons between VideoLucy and GOPAgen on LVBench, NExT-QA, and MotionBench. The displayed questions cover event counting, actions, and relative motion. Each panel pairs a question with the two model responses.}
        \label{fig:333}
        \vspace{-0.2in}
    \end{minipage}
    \hfill
    \begin{minipage}{0.48\linewidth}
        \centering
        \captionof{table}{Motion-agent substitution within the complete GOPAgen framework. Entries report accuracy (\%) on NExT-QA and LVBench, with the remaining system components retained. Bold marks the highest value among the listed alternatives. This comparison evaluates module choices, which may differ in training and model capacity, rather than isolating the effect of motion vectors alone.}
\label{tab:motion_agent_ablation}
        \renewcommand{\arraystretch}{1.25}
        \setlength{\tabcolsep}{4.5pt}
        \resizebox{\linewidth}{!}{
        \begin{threeparttable}
            \begin{tabular}{l cc}
                \toprule
                \textbf{Methods} & \textbf{NExT-QA} & \textbf{LVBench} \\
                \midrule
                \multicolumn{3}{l}{\textcolor{gray}{\textit{Open-Source VLMs as Motion Agents}}} \\
                InternVL2.5-78B~\citep{wang2025internvideo2}      & 78.7 & 70.2 \\
                VideoLLaMA3-7B~\citep{damonlpsg2025videollama3}   & 79.8 & 70.7 \\
                Qwen2.5-VL-72B~\citep{bai2025qwen2}              & 81.5 & 71.8 \\
                VideoChat-Flash~\citep{li2024videochat}          & 80.2 & 69.8 \\
                LLaVA-NeXT-Video~\citep{li2024llava}              & 80.7 & 71.3 \\
                \midrule
                \textbf{Motion Agent (Ours)} & \textbf{83.1} & \textbf{73.5} \\
                \bottomrule
            \end{tabular}
        \end{threeparttable}
        }
        \label{tab:lvbench_comparison}
    \end{minipage}
\end{figure}

We replace the motion agent with alternative video VLMs while retaining the other GOPAgen components. In~\cref{tab:motion_agent_ablation}, our agent achieves 83.1\% accuracy on NExT-QA and 73.5\% on LVBench. The strongest listed alternative achieves 81.5\% and 71.8\%, respectively. This comparison evaluates module choices within GOPAgen. It does not isolate motion-vector input from differences in training or model capacity. Supplementary experiments examine component removal and replacement in~\cref{app:component_attribution}.
\subsection{Time and Token Efficiency}
We measure latency and visual-input requirements separately because token counts alone do not determine wall-clock cost. Under the protocol in~\cref{app:preliminary_time_efficiency}, GOPAgen takes 65\,s per query, compared with 125\,s for VideoLucy and 227\,s for Deep Video Discovery (DVD). These correspond to approximately 52.0\% and 28.6\% of the respective baseline latencies. The comparison excludes one-time video preprocessing. Additional cost accounting appears in~\cref{app:end_to_end_cost}.

For a duration-based visual-input estimate, let $Ds$ denote video duration in seconds, $Rs$ the sampling rate, and $Nf$ the number of tokens per sampled frame. Under the evaluated DVD configuration, whole-video processing has the estimated visual-token count
\begin{equation}
\label{eq:777}
Num_{\text{DVD}} = Ds \times Rs \times Nf.
\end{equation}
For a 15-minute video with $Ds=900$\,s, $Rs=4$\,FPS, and $Nf=1{,}200$, this estimate is 4.32M visual tokens. The reported GOPAgen count sums visual tokens across global inspection, caption generation, and motion-description generation. It is approximately 0.061M for the corresponding single-query comparison, giving a ratio of about 1:70.8. These counts characterize the stated visual-input configurations, not an equivalent reduction in FLOPs, text tokens, or total serving cost. Multi-query use additionally depends on which preprocessing and memory-construction costs can be reused.

\section{Conclusion}
We presented GOPAgen, a codec-aware framework that integrates textual motion evidence with appearance and temporal provenance in typed memory pages. The GOP-Tree organizes these pages for conditional retrieval and final reasoning. Experiments show competitive performance across motion-focused and long-video benchmarks, supporting codec-aware structured memory for selective video understanding. Gains vary across datasets, and practical efficiency depends on preprocessing, caching, external models, and query workload. By linking appearance and motion descriptions to their source intervals, the memory interface preserves an explicit connection between retrieved evidence and the underlying video. Future work could improve temporal selection and motion descriptions to address the evidence-acquisition failures identified in our analysis.

\subsection*{AI use statement}
Generative AI tools were used solely to improve the clarity and readability
of the manuscript. They were not used to formulate scientific claims,
design experiments, generate results, or interpret findings. All
AI-assisted edits were reviewed by the authors, who take full
responsibility for the manuscript.

\subsection*{Ethics Statement}
This work does not involve human participants, personally identifiable
information, or private data. We recognize that the proposed method may
inherit biases and limitations from its training data and underlying
models. Potential misuse and unintended deployment risks should therefore
be carefully considered in real-world applications. All authors have
reviewed the ethical implications of this work and take responsibility
for its content.

\subsection*{Reproducibility Statement}
We provide detailed descriptions of the proposed method, model
configuration, training procedure, datasets, evaluation protocols, and
hyperparameters in the main paper and appendices. We report the
computational resources and random seeds used in our experiments where
applicable. The implementation details necessary to reproduce the main results are
provided in the paper and appendices.

\bibliography{references}
\bibliographystyle{iclr2027_conference}

\clearpage
\appendix
\setcounter{secnumdepth}{3}

\section{Notation}
\label{app:notation}

\noindent\begin{minipage}{\textwidth}
\centering
\small
\captionof{table}{Notation for GOPAgen and its traversal algorithm. Symbols describe video and query inputs, GOP records, selected intervals, memory pages, the GOP-Tree, and reasoning outputs. The node and edge sets define the tree; other intermediate sets identify retained nodes and pages.}
\label{tab:notation}
\begin{tabular}{p{0.23\linewidth}p{0.71\linewidth}}
\toprule
Symbol & Meaning \\
\midrule
$V$, $q$
& Input video and natural-language question. \\

$B_j$, $K_j$, $\mathrm{MV}_j$
& The $j$-th GOP record, its intra-coded reference frame, and its motion-vector records. \\

$u$, $\mathcal{B}_u$
& A selected temporal interval and the indices of GOPs overlapping that interval. \\

$m_u$, $c_u$
& Textual motion evidence and RGB-based appearance description for interval $u$. \\

$\mathcal{G}$
& Coarse, query-conditioned textual memory of the complete video. \\

$\mathcal{U}_{c}$, $\mathcal{U}_{f}$
& Retained coarse intervals and retained fine intervals after the two selection rounds. \\

$N_g$, $k_1$, $k_2$
& Numbers of globally sampled frames, retained coarse intervals, and retained fine intervals, respectively. \\

$P_u$
& Typed memory page associated with interval $u$. \\

$z_u$, $\pi_u$
& Retrieval embedding and provenance record stored in page $P_u$. \\

$\mathcal{M}_q$
& Query-conditioned collection of typed memory pages. \\

$\mathcal{T}$, $\mathcal{N}$, $\mathcal{E}$
& GOP-Tree and its node and edge sets. \\

$A_s$, $A_r$
& Local ranker and final reasoning agent, used explicitly in the Appendix pseudocode. \\

$\mathcal{N}_1$, $\mathcal{N}_2$, $\mathcal{P}$, $\mathcal{P}_{\mathrm{rank}}$
& Retained nodes at the two tree levels, their resolved pages, and the ranked page subset in the traversal algorithm. \\

$\mathcal{G}_{\mathrm{cmp}}$, $b$, $\mathcal{C}$, $o$
& Compressed global memory, sufficiency decision, optional local context, and final answer in the traversal algorithm. \\
\bottomrule
\end{tabular}
\end{minipage}
\par\medskip

\section{Implementation and Reproducibility Details}
\label{app:implementation_details}

\subsection{Motion-Agent Training}
\label{app:motion_training_details}
Training uses three successive datasets: 558,000 image--text alignment pairs, 1.2 million mixed image/video instruction pairs, and 20,000 LLaVA-Hound motion-caption examples. Their sizes have an approximate ratio of 31:68:1; this describes dataset composition rather than exposure across epochs.

All stages use AdamW, cosine learning-rate decay, 3\% warm-up, bfloat16 precision, gradient clipping at 1.0, and seed 42. Pretraining runs for one epoch with learning rate $1\times10^{-3}$, zero weight decay, effective batch size 256, and 336-pixel inputs. Mid-training runs for one epoch with learning rate $2\times10^{-5}$, weight decay 0.01, effective batch size 128, and 448-pixel inputs. Motion fine-tuning runs for three epochs with learning rate $1\times10^{-5}$, weight decay 0.05, effective batch size 128, and 448-pixel inputs. All stages use four gradient-accumulation steps.

Training runs on eight NVIDIA A100 80-GB GPUs. The stages take approximately 4.5, 9, and 2.25 hours, corresponding to 36, 72, and 18 GPU-hours. During pretraining, the vision encoder and language model are frozen, while the multimodal projector and motion tokenizer are optimized. During mid-training, the projector, motion tokenizer, and language model are trainable; the vision encoder remains frozen. Motion fine-tuning optimizes the motion tokenizer, projector, and rank-16 LoRA adapters in the language model. Corrupted samples are removed, near-duplicates are filtered using perceptual hashes, and overlength sequences are uniformly subsampled.

\subsection{Codec Extraction and Motion Tokenization}
\label{app:codec_details}
Videos are standardized to H.264/AVC High Profile at 1280$\times$720 resolution and 30 FPS. Encoding uses a closed GOP of 60 frames, two B-frames, CRF 23, and disabled scene-cut insertion. The FFmpeg 6.1 codec options are \texttt{-c:v libx264 -profile:v high -crf 23 -g 60 -keyint\_min 60 -bf 2 -sc\_threshold 0 -pix\_fmt yuv420p}. Motion vectors are exported with \texttt{-flags2 +export\_mvs} and read through PyAV 12.0.0.

I/IDR frames supply RGB reference images, and valid motion records from P- and B-frames are retained. Reference direction distinguishes predictions from past and future frames; it is distinct from an exact reference-picture identifier. For width $W$ and height $H$, block centers and pixel displacements are normalized as $(x/W,y/H,\Delta x/W,\Delta y/H)$ and clipped to $[-1,1]$. Here, $\Delta x$ and $\Delta y$ denote displacements after accounting for the motion-vector scale. Invalid records are discarded, and blocks without vectors receive a zero vector and a missing-vector mask. Records are ordered by timestamp and then raster position.

Each token includes position, displacement, block size, temporal offset, reference direction, and validity state. A four-layer spatiotemporal Transformer maps a $16\times32\times32$ grid to 1,024-dimensional features. At most 512 motion tokens are retained per GOP and 8,192 per query. Longer sequences are uniformly subsampled, while shorter sequences are padded with an attention mask. For intervals spanning multiple GOPs, overlapping GOP records are loaded and restricted to the selected timestamps.

\subsection{Global Sampling and Temporal Selection}
\label{app:selection_details}
Global inspection samples one frame per second, with at most $N_g=32$ frames. Videos exceeding this cap are uniformly resampled over their full duration. Frames use the Qwen2.5-VL processor at 448$\times$448 resolution. At a spatial patch size of 14, each frame contains 1,024 patches before spatial merging, or 32,768 patches across 32 frames. These patch counts are distinct from the visual tokens supplied to the language model.

The timeline is divided into non-overlapping 60-second coarse candidates, of which $k_1=3$ are retained. Each retained interval is partitioned into 8-second fine candidates, and the second round retains $k_2=2$ candidates per coarse interval. The nominal default retains at most six fine intervals and covers 48 seconds before GOP expansion. Candidate inspection counts and GOP-expanded processing durations are reported separately from this retention budget. Qwen2.5-VL scores relevance using the question, timestamps, and global summary with the instruction \textit{\textquotedblleft Return a relevance score from 0 to 100 and one supporting sentence for each candidate interval.\textquotedblright} Scores are parsed as integers in $[0,100]$. Malformed outputs are retried once and then assigned zero; ties favor the earlier timestamp. Fine intervals with temporal intersection-over-union above 0.5 are merged, retaining the highest score.

\subsection{Memory Page and Index Configuration}
\label{app:index_details}
Each memory page corresponds to a retained fine interval after overlap handling. Retrieval embeddings are computed using \texttt{BAAI/bge-large-en-v1.5}, mean pooling, and $\ell_2$ normalization, yielding 1,024-dimensional vectors. ChromaDB 0.5.23 stores the page identifier, embedding, appearance and motion descriptions, video identifier, parent identifier, timestamps, GOP identifiers, model names, and prompt hashes.

RGB frames and motion-vector arrays are stored externally as compressed JPEG and NPZ objects referenced by stable identifiers. The persistent index uses cosine distance and HNSW with $M=16$, \texttt{efConstruction}=200, and \texttt{efSearch}=64. SQLite metadata and HNSW files provide local persistence. Video identity and temporal overlap determine page eligibility before ranking.

Pages are inserted in batches of 64 with deterministic identifiers. Repeated identifiers are ignored unless overwrite is enabled, and duplicate video/timestamp ranges are rejected. Cold-cache measurements restart the process and clear the operating-system file cache; warm-cache measurements reuse the initialized collection.

\subsection{Conditional Reasoning and Traversal}
\label{app:reasoning_details}
Before local-agent calls and page construction, DeepSeek-V3 assesses global sufficiency with temperature 0 and at most 256 output tokens. The prompt requests JSON fields \texttt{sufficient}, \texttt{confidence}, and \texttt{rationale}. A positive decision requires \texttt{sufficient=true} and confidence of at least 0.70, selected on a 500-question development subset.

The local ranker, \texttt{BAAI/bge-reranker-v2-m3}, scores candidates with an FP16 cross-encoder. Traversal retains three coarse nodes, two fine nodes per coarse node, and at most five final pages. DeepSeek-V3 compresses the global memory and retrieved evidence using a 16,000-token input cap and a 1,024-token output cap. Evidence sharing an interval and having cosine text similarity above 0.90 is deduplicated before compression.

Final reasoning uses temperature 0, a 64-token output limit, a 60-second timeout, and at most three retries. Evaluation uses answer matching rather than model-based judging. A regular expression extracts the first valid option label in \texttt{A--E} and compares it with the ground-truth option. Outputs without a valid label are counted as invalid. API experiments used the \texttt{deepseek-chat} endpoint accessed on 15 August 2026.

\begin{algorithm}[t]
\caption{Global-sufficiency gating followed by on-demand page construction and GOP-Tree retrieval. BuildPages runs the caption and motion agents; RankSelect applies the budgets in~\cref{app:reasoning_details}.}
\label{alg:gop_tree_reasoning}
\KwIn{Video $V$, question $q$, global memory $\mathcal{G}$, selected intervals $\mathcal{U}_{c},\mathcal{U}_{f}$, ranker $A_s$, reasoning agent $A_r$}
\KwOut{Answer $o$}
$\mathcal{G}_{\mathrm{cmp}}\leftarrow\operatorname{Compress}(\mathcal{G})$\;
$b\leftarrow\operatorname{Sufficient}(A_r,q,\mathcal{G}_{\mathrm{cmp}})$\;
\eIf{$b=\mathrm{true}$}{
    $\mathcal{C}\leftarrow\varnothing$\;
}{
    $\mathcal{M}_q\leftarrow\operatorname{BuildPages}(V,\mathcal{U}_{f})$\;
    $\mathcal{T}\leftarrow\operatorname{BuildTree}(\mathcal{G},\mathcal{U}_{c},\mathcal{U}_{f},\mathcal{M}_q)$\;
    $\mathcal{N}_1\leftarrow\operatorname{RankSelect}(A_s,\operatorname{children}(\operatorname{root}(\mathcal{T})),q)$\;
    $\mathcal{N}_2\leftarrow\varnothing$\;
    \ForEach{$v\in\mathcal{N}_1$}{
        $\mathcal{N}_2\leftarrow\mathcal{N}_2\cup\operatorname{RankSelect}(A_s,\operatorname{children}(v),q)$\;
    }
    $\mathcal{P}\leftarrow\operatorname{ResolvePages}(\mathcal{N}_2,\mathcal{M}_q)$\;
    $\mathcal{P}_{\mathrm{rank}}\leftarrow\operatorname{RankSelect}(A_s,\mathcal{P},q)$\;
    $\mathcal{C}\leftarrow\operatorname{Compress}(\mathcal{P}_{\mathrm{rank}})$\;
}
$o\leftarrow A_r(\operatorname{Template}(q,\mathcal{G}_{\mathrm{cmp}},\mathcal{C}))$\;
\Return{$o$}\;
\end{algorithm}

\section{Agent Prompts and Runtime Details}
\label{app:agent_prompts_runtime}
In this section, we provide additional implementation details for GOPAgen.

\subsection{Implementation Prompts}
\label{app:implementation_prompts}
We provide the instruction templates and configurations used by each agent.

\subsubsection{Global Agent}
\label{app:global_agent_prompt}
For the global agent, the instruction template is: ``Please provide as much detail shown in the video about the following question as possible:''. We adopt the system-prompt configuration of Qwen2.5-VL, where the \texttt{qwen\_chat} dictionary is set as follows:
\begin{PromptBlock}
qwen_chat=dict(
    SYSTEM=('<|im_start|>system\n{system}<|im_end|>\n'),
    INSTRUCTION=('<|im_start|>user\n{input}<|im_end|>\n'
                 '<|im_start|>assistant\n'),
    SUFFIX='<|im_end|>',
    SUFFIX_AS_EOS=True,
    SEP='\n',
    STOP_WORDS=['<|im_end|>', '<|endoftext|>'])
\end{PromptBlock}

\subsubsection{Caption Agent}
\label{app:caption_agent_prompt}
The instruction template for the detailed captioning agent is:
\begin{PromptBlock}
"The video lasts for {video_time:.2f} seconds, and {len(video[0])} frames are uniformly" + \
"sampled from it. These frames are located at {frame_time}. Please provide detailed " + \
"caption of this video."
\end{PromptBlock}
Here, \texttt{video\_time:.2f} denotes the video duration rounded to two decimal places, \texttt{len(video[0])} is the number of uniformly sampled frames, and \texttt{frame\_time} contains their timestamps.

\subsubsection{Motion Agent}
\label{app:motion_agent_prompt}
The instruction template for the motion agent is:
\begin{PromptBlock}
"The video lasts for {video_time:.2f} seconds, and {len(video[0])} frames are uniformly" +\
"sampled from it. These frames are located at {frame_time}. Please" +\
"provide detailed motion description of this video."
\end{PromptBlock}
The parameters have the same meanings as those in the detailed captioning-agent template.

\subsubsection{Long-Context Reasoning Agent}
\label{app:reasoning_agent_prompt}
The prompt for the long-context reasoning agent, denoted by \texttt{memory\_prompt}, is constructed as follows:
\begin{PromptBlock}
memory_prompt = "Please read the following video content descriptions and the question " +\
"in depth. The video content is presented in a coarse-to-fine way and more motion details are" +\
"included when zooming in.\n"
memory_prompt = memory_prompt + "Coarse Memory: "+str(root_memory) + "\n"
memory_prompt = memory_prompt + "Fine Memory: "+str(subtree_memory) + "\n"
memory_prompt = memory_prompt + "Now, a question has been raised regarding the content " + \
                "descriptions of this video.\n" + question + "\n"
memory_prompt = memory_prompt + "Please combine all the content analysis above together and" + \
                "provide a final answer to the question: "  + question + " If the provided" + \
                "context is inconsistent, insufficient, or contradictory, please adopt" + \
                "the motion agent's reference answer as the final response." + "\n" +\
                "Please answer in the following format: Final answer: A."
\end{PromptBlock}
Here, \texttt{root\_memory} is the coarse-grained memory, \texttt{subtree\_memory} is the fine-grained memory, and \texttt{question} is the target question. The template above reproduces the wording used in the experiments. Its reference to a motion-agent answer is not a separate output field: the motion agent supplies only fine-grained motion descriptions. The final answer is generated by the reasoning agent from the supplied evidence. The caption agent contributes the appearance description for each selected interval; the motion agent supplies the complementary motion description.

\subsection{Motion-Vector Database Scheduling}
\label{app:mv_database_scheduling}
Each extracted GOP record receives a stable video/GOP identifier and temporal interval $[s_j,e_j]$. Records reference motion-vector arrays and store codec, reference-direction, block-coordinate, displacement, and timestamp metadata. The page collection associates $z_u$ from~\cref{eq:memory_page} with its selected interval and parent identifier. Retrieval first filters overlapping intervals and then resolves the referenced arrays.

The database uses the persistence, cosine-distance HNSW index, insertion batches, and duplicate policy described in~\cref{app:index_details}. Raw motion arrays remain in external NPZ storage. Table~\ref{tab:e2e_cost} separates construction, database access, model computation, and storage costs.

\subsection{Latency Comparison}
\label{app:preliminary_time_efficiency}
Figure~\ref{fig:latency} compares average processing times of 65\,s for GOPAgen, 125\,s for VideoLucy, and 227\,s for Deep Video Discovery. The comparison uses 200 LVBench validation videos spanning 10--60 minutes, stratified by duration, with one question per video and three runs per method. The corresponding standard deviations are 4.8\,s, 9.7\,s, and 18.5\,s.

All methods process one query at a time on a server with one NVIDIA A100 80-GB GPU and 64 CPU cores. Latency uses a cold application cache and includes model inference, retrieval, and API waiting, but excludes one-time video preprocessing. API calls used the \texttt{deepseek-chat} endpoint accessed on 15 August 2026. These measurements characterize this workload; the stage-wise accounting in~\cref{app:end_to_end_cost} separately reports preprocessing and cache-dependent costs.

\section{Supplementary Experiments}
\label{app:supplementary_experiments}
The supplementary experiments examine motion representation, temporal organization, retrieval behavior, and computational cost. Compared variants share benchmark questions, answer parsing, agent models, prompts, and decoding settings unless the corresponding experiment changes those components. Component-specific results are interpreted alongside the realized inference budgets.

\subsection{Common Evaluation Protocol}
\label{app:common_protocol}
Controlled variants are evaluated on fixed 500-question subsets of MotionBench Test and LVBench validation, sampled across duration and question-type strata. Each stochastic or API-based experiment is repeated three times, and tables report mean results. Uncertainty is assessed using question-level 95\% bootstrap confidence intervals with 10,000 paired resamples. Pairwise accuracy differences use two-sided paired permutation tests with 10,000 permutations.

Runs use one NVIDIA A100 80-GB GPU, 64 CPU cores, CUDA 12.1, PyTorch 2.3, and bfloat16 model inference unless otherwise specified. Checkpoints are fixed, queries are processed sequentially, and API calls have a 60-second timeout with at most three retries. The reasoning endpoint was \texttt{deepseek-chat}, accessed on 15 August 2026. Accuracy experiments use a warm model cache; resource measurements distinguish cold and warm caches.

For component comparisons, we match evidence and inference budgets where the representations permit. We report realized duration, token counts, and model calls when they differ. Consequently, replacing an entire model or changing its input representation is not interpreted as a compute-identical intervention.

\subsection{Controlled Component Attribution}
\label{app:component_attribution}
The component comparison varies motion input, temporal organization, and memory structure. \emph{No MV} removes codec motion vectors while retaining reference frames and the memory schema. \emph{Reference-frame only} retains the motion-agent call but supplies RGB reference frames alone. \emph{Fixed windows} replaces GOP-based organization with fixed-duration intervals at the reported evidence budget. \emph{Flat memory} preserves page content but removes parent--child containment edges. \emph{Optical flow} replaces codec vectors with RAFT-Large flow computed on 720p frame pairs. Flow is normalized by frame dimensions, rasterized to the same $16\times32\times32$ grid, and limited to the same 8,192-token motion budget.

Table~\ref{tab:component_ablation} reports both accuracy and realized resource use. The full system achieves higher accuracy than the listed variants on MotionBench Test and LVBench. Because some interventions also change token counts or model calls, their effects reflect the complete stated configuration.

\begin{table*}[t]
\centering
\color{black}
\caption{Component comparison on MotionBench Test and LVBench. Accuracy is reported in percent. Selected duration is the temporal evidence processed per query; token counts and model calls describe realized inference budgets. Latency is wall-clock time per query. Rows remove or replace motion input, temporal organization, or memory structure. Resource differences are reported explicitly rather than treated as compute-identical comparisons.}
\label{tab:component_ablation}
\resizebox{\linewidth}{!}{
\begin{tabular}{lccccccc}
\toprule
Variant & MotionBench & LVBench & Selected duration & Visual tokens & Motion tokens & Calls/query & Latency/query \\
\midrule
Full GOPAgen & 65.3 & 73.5 & 48 s & 61k & 8k & 9 & 65.0 s \\
No motion-vector input & 60.8 & 70.1 & 48 s & 61k & 0 & 9 & 59.4 s \\
Reference-frame only & 61.9 & 70.8 & 48 s & 61k & 0 & 9 & 60.6 s \\
Caption only (no motion agent) & 59.7 & 69.4 & 48 s & 61k & 0 & 7 & 48.9 s \\
Fixed windows (no GOP boundaries) & 63.1 & 71.6 & 48 s & 63k & 8k & 9 & 66.8 s \\
Flat memory (no GOP-Tree) & 62.7 & 71.9 & 48 s & 61k & 8k & 9 & 63.7 s \\
Optical flow instead of codec MV & 64.1 & 72.6 & 48 s & 61k & 8k & 9 & 83.2 s \\
\bottomrule
\end{tabular}}
\end{table*}

\subsection{Agentic Baseline Comparison}
\label{app:matched_baselines}
We compare agentic baselines on the same videos and questions using a shared answer parser, final reasoning endpoint, hardware, precision, and cache policy. Each method retains its evidence-acquisition procedure, so visual-token counts and model-call counts can differ. Table~\ref{tab:matched_baselines} therefore reports accuracy alongside realized resource use rather than assuming equal inference budgets.

\begin{table*}[t]
\centering
\color{black}
\caption{Accuracy and resource use for agentic baselines on LVBench under shared hardware and final-reasoner settings. All methods use DeepSeek-V3 for final reasoning, while evidence acquisition and model-call counts follow each method's configuration. Visual tokens and calls are reported per query, accuracy in percent, and latency in seconds per query. API cost is normalized to GOPAgen ($1.0\times$); equal hardware does not imply equal inference budgets.}
\label{tab:matched_baselines}
\resizebox{\linewidth}{!}{
\begin{tabular}{lcccccc}
\toprule
Method & Reasoner & Visual tokens/query & Calls/query & Accuracy & Latency/query & Relative API cost \\
\midrule
VideoTree & DeepSeek-V3 & 0.180M & 18 & 28.8 & 142 s & 2.8$\times$ \\
VideoLucy & DeepSeek-V3 & 0.240M & 22 & 58.8 & 125 s & 3.5$\times$ \\
Deep Video Discovery & DeepSeek-V3 & 4.320M & 32 & 74.2 & 227 s & 8.7$\times$ \\
GOPAgen & DeepSeek-V3 & 0.061M & 9 & 73.5 & 65 s & 1.0$\times$ \\
\bottomrule
\end{tabular}}
\end{table*}

\subsection{End-to-End Cost Accounting}
\label{app:end_to_end_cost}
We distinguish one-time video preprocessing from online query processing. Measurements include decoding, data movement, serialization, database operations, model inference, and network/API time. Resource measurements use 50 videos in each of three duration bins: 5--15, 15--30, and 30--60 minutes. Three questions per video yield 150 videos and 450 query measurements. Table~\ref{tab:e2e_cost} provides stage-wise accounting for a 15-minute video and distinguishes cold-cache, warm-cache, and amortized costs.

\begin{table*}[t]
\centering
\color{black}
\caption{Stage-wise resource accounting for a 15-minute video. The first block reports video-level preprocessing and construction costs; the second reports online processing per query. API tokens are input/output counts, GPU memory is the peak within each stage, and storage covers motion arrays, page metadata, and index files. Cold-cache and warm-cache online totals are shown separately from the ten-query amortized estimate.}
\label{tab:e2e_cost}
\resizebox{\linewidth}{!}{
\begin{tabular}{lrrrrrrr}
\toprule
Stage & Wall time & GPU time & CPU time & Peak GPU memory & API input/output tokens & API cost & Storage \\
\midrule
Decode + MV extraction (per video) & 14.2 s & 0.0 s & 13.8 s & 1.2 GB & 0 & 0 & 48.0 MB \\
Global frame sampling (per video) & 2.8 s & 1.9 s & 0.9 s & 3.4 GB & 0 & 0 & 12.0 MB \\
Page/index construction (per video) & 6.5 s & 1.7 s & 4.8 s & 4.1 GB & 0/0 & \$0.000 & 3.6 MB \\
\midrule
Global inspection (per query) & 9.8 s & 8.7 s & 1.1 s & 21.6 GB & 9.6k/0.9k & \$0.006 & -- \\
Two-stage temporal selection (per query) & 2.6 s & 1.8 s & 0.8 s & 18.2 GB & 5.2k/0.4k & \$0.003 & -- \\
Caption-agent calls (per query) & 16.4 s & 14.9 s & 1.5 s & 22.4 GB & 0/0 & \$0.000 & -- \\
Motion-agent calls (per query) & 20.7 s & 18.6 s & 2.1 s & 23.8 GB & 0/0 & \$0.000 & -- \\
Database query/update (per query) & 1.3 s & 0.0 s & 1.2 s & 0.8 GB & 0 & 0 & -- \\
Ranking and compression (per query) & 3.2 s & 1.4 s & 1.8 s & 7.6 GB & 2.8k/0.3k & \$0.002 & -- \\
Final reasoning (per query) & 11.0 s & 0.0 s & 0.7 s & 1.1 GB & 18.4k/0.3k & \$0.011 & -- \\
\midrule
Online total, cold cache & 65.0 s & 45.4 s & 9.2 s & 23.8 GB & 36.0k/1.9k & \$0.022 & -- \\
Online total, warm cache & 58.4 s & 43.8 s & 6.4 s & 23.8 GB & 36.0k/1.9k & \$0.022 & -- \\
Amortized total over $Q{=}$10 queries & 60.8 s/query & 44.0 s/query & 8.0 s/query & 23.8 GB & 36.0k/1.9k & \$0.022/query & 63.6 MB/video \\
\bottomrule
\end{tabular}}
\end{table*}

\subsection{Temporal-Selection Sensitivity}
\label{app:selection_sensitivity}
We vary the stated temporal-selection parameter while retaining the remaining nominal settings. Table~\ref{tab:selection_sensitivity} reports accuracy, GOP-expanded processing duration, inspected candidates, token counts, and latency. Candidate counts include examined candidates, not only the final retained intervals or retrieved pages. Processing duration includes GOP expansion and can exceed the nominal selected-interval duration. Reporting these quantities exposes changes in realized evidence and computation rather than treating every configuration as cost-equivalent.

\begin{table*}[t]
\centering
\color{black}
\caption{Sensitivity to global sampling, temporal selection, reference-frame policy, and motion-token budget on LVBench. Slash-separated values follow the three settings in each row; bold identifies the default. Accuracy is in percent and token counts in thousands. GOP-expanded duration reports processed evidence, not only nominal selected intervals. Candidates/query includes inspected candidates rather than only retained pages; it can exceed the final retention budget. Duration and latency are in seconds.}
\label{tab:selection_sensitivity}
\resizebox{\linewidth}{!}{
\begin{tabular}{llcccccc}
\toprule
Factor & Setting & Accuracy & GOP-expanded duration & Candidates/query & Visual tokens & Motion tokens & Latency/query \\
\midrule
Global sampling & 0.5/\textbf{1}/2 FPS & 71.8/\textbf{73.5}/73.7 & 48/\textbf{48}/48 s & 4.6/\textbf{4.8}/4.9 & 34/\textbf{61}/118k & 8/\textbf{8}/8k & 53/\textbf{65}/91 s \\
Coarse interval & 30/\textbf{60}/120 s & 72.6/\textbf{73.5}/72.9 & 45/\textbf{48}/54 s & 5.2/\textbf{4.8}/4.1 & 59/\textbf{61}/65k & 8/\textbf{8}/9k & 67/\textbf{65}/64 s \\
Coarse top-$k_1$ & 2/\textbf{3}/4 & 72.1/\textbf{73.5}/73.8 & 32/\textbf{48}/64 s & 3.3/\textbf{4.8}/6.2 & 43/\textbf{61}/79k & 6/\textbf{8}/11k & 51/\textbf{65}/82 s \\
Fine interval & 4/\textbf{8}/16 s & 72.8/\textbf{73.5}/72.7 & 40/\textbf{48}/64 s & 7.8/\textbf{4.8}/3.2 & 58/\textbf{61}/68k & 9/\textbf{8}/8k & 73/\textbf{65}/61 s \\
Fine top-$k_2$ & 1/\textbf{2}/3 & 71.9/\textbf{73.5}/73.9 & 24/\textbf{48}/72 s & 2.5/\textbf{4.8}/7.1 & 39/\textbf{61}/86k & 4/\textbf{8}/12k & 47/\textbf{65}/88 s \\
Reference-frame policy & IDR/\textbf{I+IDR}/all refs & 72.4/\textbf{73.5}/73.3 & 48/\textbf{48}/48 s & 4.8/\textbf{4.8}/4.8 & 57/\textbf{61}/69k & 6/\textbf{8}/11k & 61/\textbf{65}/74 s \\
Motion-token budget & 4/\textbf{8}/16k & 72.2/\textbf{73.5}/73.8 & 48/\textbf{48}/48 s & 4.8/\textbf{4.8}/4.8 & 61/\textbf{61}/61k & 4/\textbf{8}/16k & 59/\textbf{65}/78 s \\
\bottomrule
\end{tabular}}
\end{table*}

\subsection{GOP-Tree Retrieval Diagnostics}
\label{app:retrieval_diagnostics}
Retrieval is evaluated on 300 manually annotated LVBench validation questions spanning question types and video-duration bins. A page is relevant when its temporal intersection-over-union with an annotated evidence interval is at least 0.30. Three annotators independently label whether global evidence is sufficient. Majority vote gives the initial label; non-unanimous cases are resolved through joint adjudication. We also compare appearance-only, motion-only, and joint page representations under a five-page retrieval budget. The mean of 4.8 retrieved pages is conditioned on queries that enter the local-retrieval path. Global-only queries, which constitute 42.6\% of the evaluation set, do not contribute to this conditional mean. The default fine-interval retention budget is six; the reported 6.2 counts examined fine-node candidates rather than retained nodes.

\begin{table}[t]
\centering
\color{black}
\caption{GOP-Tree diagnostics on 300 annotated LVBench validation questions. Precision, recall, and F1 evaluate global sufficiency. The default fine-interval retention budget is six, whereas the fine-node count includes examined candidates. The 4.8-page mean is computed only over queries entering local retrieval; global-only queries are excluded. Evidence recall@5 uses temporal intersection-over-union of at least 0.30. Compression and representation comparisons use the stated five-page retrieval budget.}
\label{tab:retrieval_diagnostics}
\resizebox{\linewidth}{!}{
\begin{tabular}{lc}
\toprule
Diagnostic & Result \\
\midrule
Queries answered by global-only path (\%) & 42.6 \\
Sufficiency-judge precision / recall / F1 & 82.1 / 76.4 / 79.1 \\
Coarse nodes visited per query & 3.0 \\
Fine-node candidates examined per query & 6.2 \\
Pages retrieved per local-retrieval query & 4.8 \\
Evidence recall@5 & 86.7 \\
Accuracy before page compression & 72.9 \\
Accuracy after page compression & 73.5 \\
Appearance-only / motion-only / joint retrieval & 70.4 / 68.9 / 73.5 \\
\bottomrule
\end{tabular}}
\end{table}

\subsection{Motion Representation and Codec Robustness}
\label{app:codec_robustness}
Codec motion vectors depend on the encoder and do not represent optical flow. We compare H.264 High Profile with H.265/HEVC Main Profile using the same source videos, resolution, frame rate, GOP length, B-frame count, and CRF setting. Benchmark questions and labels are unchanged. This comparison evaluates the resulting encoding configurations, rather than assuming identical decoded pixels or equivalent quality at the same CRF.

We additionally exclude B-frame vectors or replace codec vectors with RAFT-Large optical flow. Table~\ref{tab:codec_robustness} reports overall accuracy and results for scene cuts, occlusion, small-object motion, and camera motion.

\begin{table*}[t]
\centering
\color{black}
\caption{Motion-representation and codec comparisons on MotionBench Test. Rows compare the default H.264 configuration with H.265 encoding, removal of B-frame vectors, and optical flow under the stated motion-token budget. Overall and condition-specific columns report accuracy (\%); latency is wall-clock time per query. Codec variants use the same source videos and benchmark labels.}
\label{tab:codec_robustness}
\resizebox{\linewidth}{!}{
\begin{tabular}{lcccccc}
\toprule
Setting & Overall accuracy & Scene cuts & Occlusion & Small objects & Camera motion & Latency/query \\
\midrule
Default codec/GOP configuration & 65.3 & 61.2 & 63.4 & 60.8 & 62.1 & 65.0 s \\
Alternative codec/GOP configuration & 64.9 & 60.6 & 63.0 & 60.1 & 61.8 & 68.4 s \\
Without B-frame vectors & 64.1 & 59.8 & 61.7 & 59.4 & 61.3 & 60.9 s \\
Optical flow under matched budget & 64.6 & 61.9 & 63.8 & 60.5 & 62.4 & 83.2 s \\
\bottomrule
\end{tabular}}
\end{table*}

\subsection{Failure Analysis}
\label{app:failure_analysis}
We categorize 100 incorrect predictions: 50 sampled from MotionBench Test and 50 from LVBench validation, using seed 42. Each example is assigned to the earliest identifiable failure in temporal selection, sufficiency assessment, motion description, appearance description, page ranking/compression, or final reasoning. Two annotators label each case independently, and a third resolves disagreements. Table~\ref{tab:failure_analysis} separates evidence-acquisition failures from errors made despite available evidence.

\begin{table}[t]
\centering
\color{black}
\caption{Failure categories for 100 incorrect predictions: 50 from MotionBench Test and 50 from LVBench validation. Each case is assigned to its earliest identifiable pipeline failure. Counts and percentages coincide because the sample contains 100 errors; categories are mutually exclusive and sum to 100. Example IDs identify representative cases. These proportions describe sampled errors, not error rates over all evaluated questions.}
\label{tab:failure_analysis}
\resizebox{\linewidth}{!}{
\begin{tabular}{lcc}
\toprule
Failure category & Count (\%) & Example IDs \\
\midrule
Relevant interval not selected & 24 (24\%) & E03, E17, E42 \\
Incorrect global-sufficiency decision & 11 (11\%) & E08, E31, E76 \\
Incorrect or incomplete motion description & 21 (21\%) & E12, E44, E69 \\
Incorrect or incomplete appearance description & 14 (14\%) & E06, E53, E88 \\
Page-ranking or compression failure & 13 (13\%) & E21, E57, E91 \\
Final reasoning failure despite sufficient evidence & 17 (17\%) & E28, E64, E97 \\
\bottomrule
\end{tabular}}
\end{table}

\section{Additional Visualizations}
\label{app:visualization_results}
\begin{figure}[t]
    \centering
    \includegraphics[width=\linewidth]{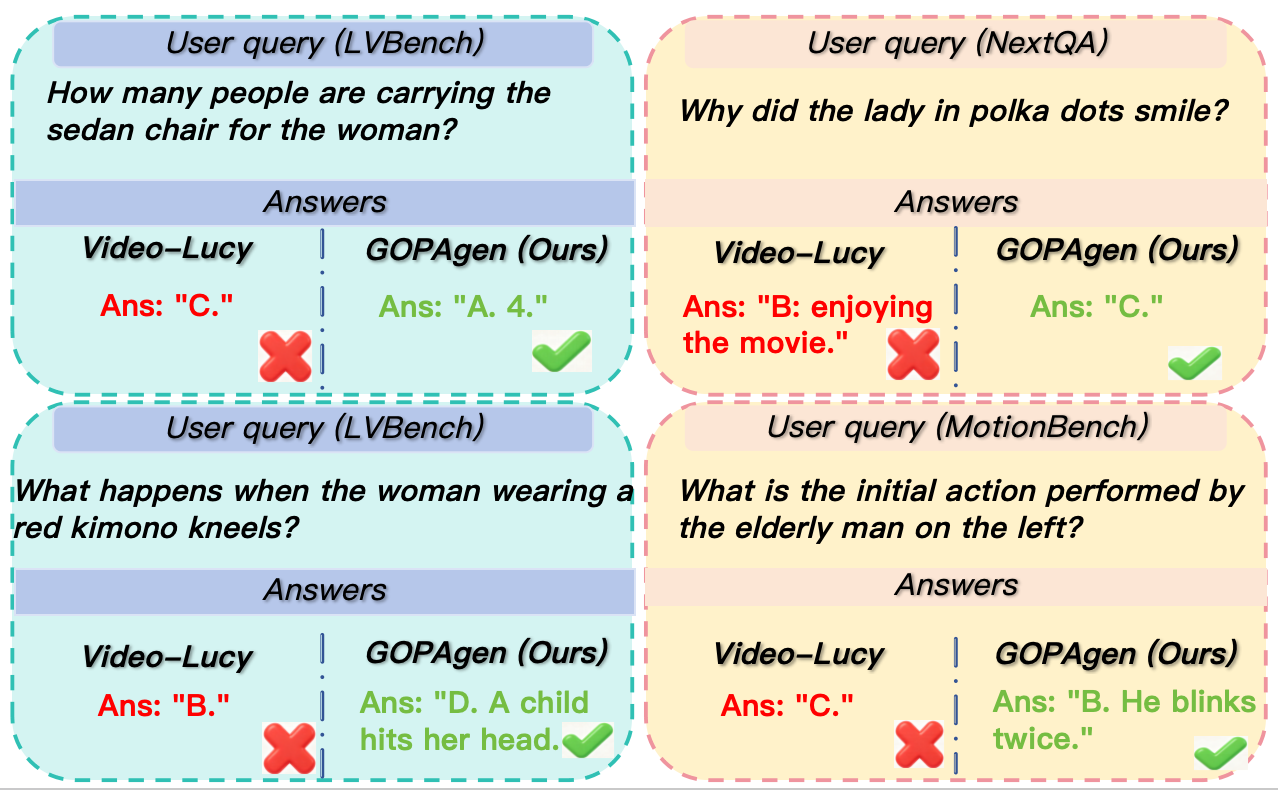}
    \caption{Selected answer comparisons with VideoLucy on LVBench, NExT-QA, and MotionBench. Questions concern the number of sedan-chair carriers, a character's reaction, a kneeling event, and an elderly man's initial action. Green check marks and red crosses indicate correct and incorrect responses in the displayed cases, respectively.}
    \label{fig:app1}
\end{figure}

\begin{figure}[t]
    \centering
    \includegraphics[width=\linewidth]{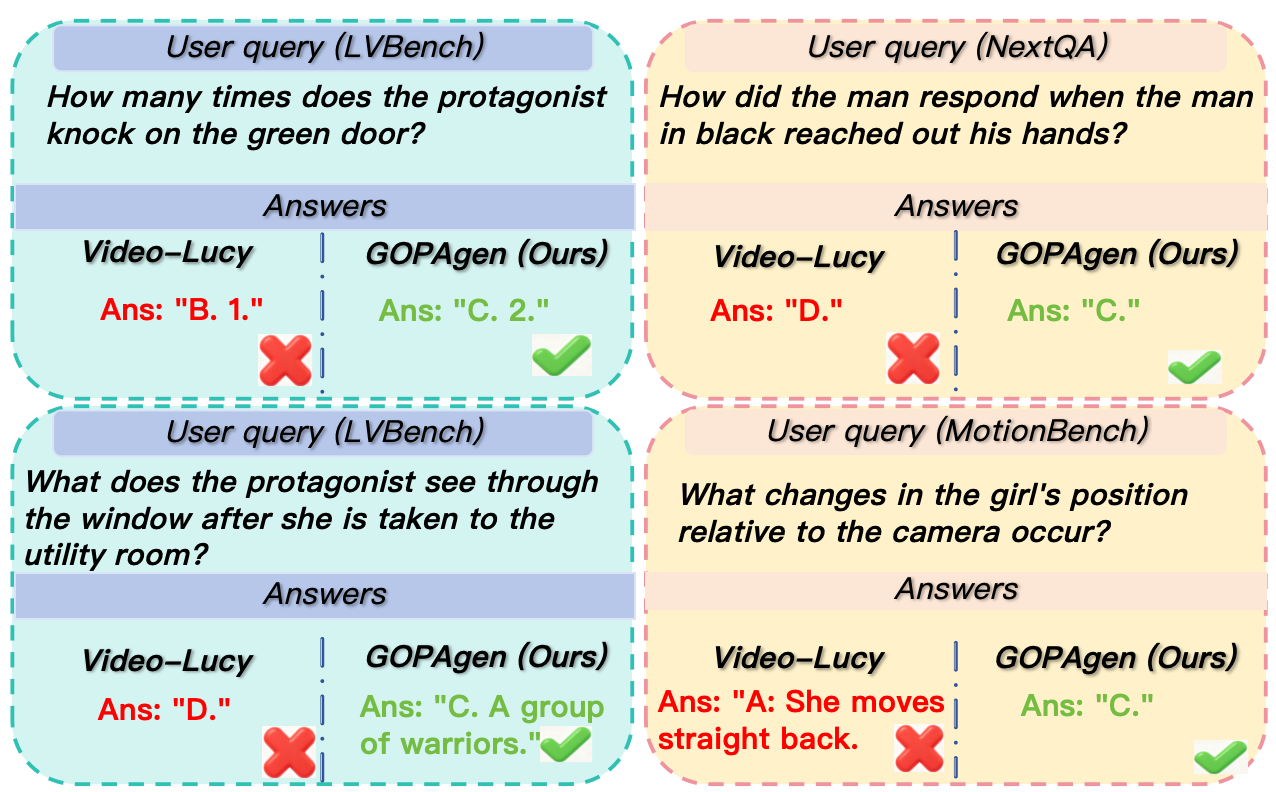}
    \caption{Additional answer comparisons with VideoLucy on LVBench, NExT-QA, and MotionBench. The examples cover counting door knocks, a response to an outstretched hand, identifying a view through a window, and motion relative to the camera. Correctness markers apply to the individual displayed responses.}
    \label{fig:app2}
\end{figure}

\begin{figure}[t]
    \centering
    \includegraphics[width=\linewidth]{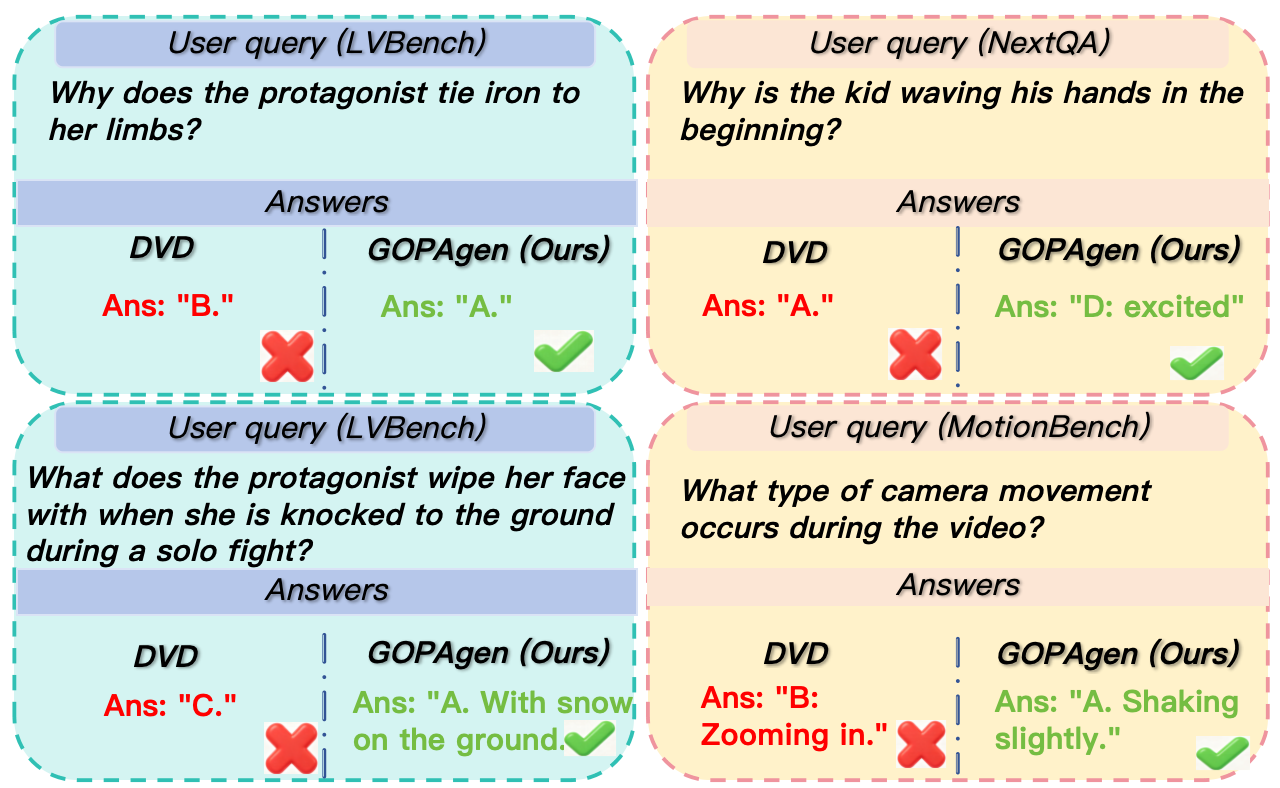}
    \caption{Selected answer comparisons with Deep Video Discovery (DVD) on LVBench, NExT-QA, and MotionBench. The questions concern a character's motivation, a child's gesture, an object used during a fight, and camera movement. Green checks mark correct responses and red crosses mark incorrect responses in these cases.}
    \label{fig:app3}
\end{figure}

\begin{figure}[t]
    \centering
    \includegraphics[width=\linewidth]{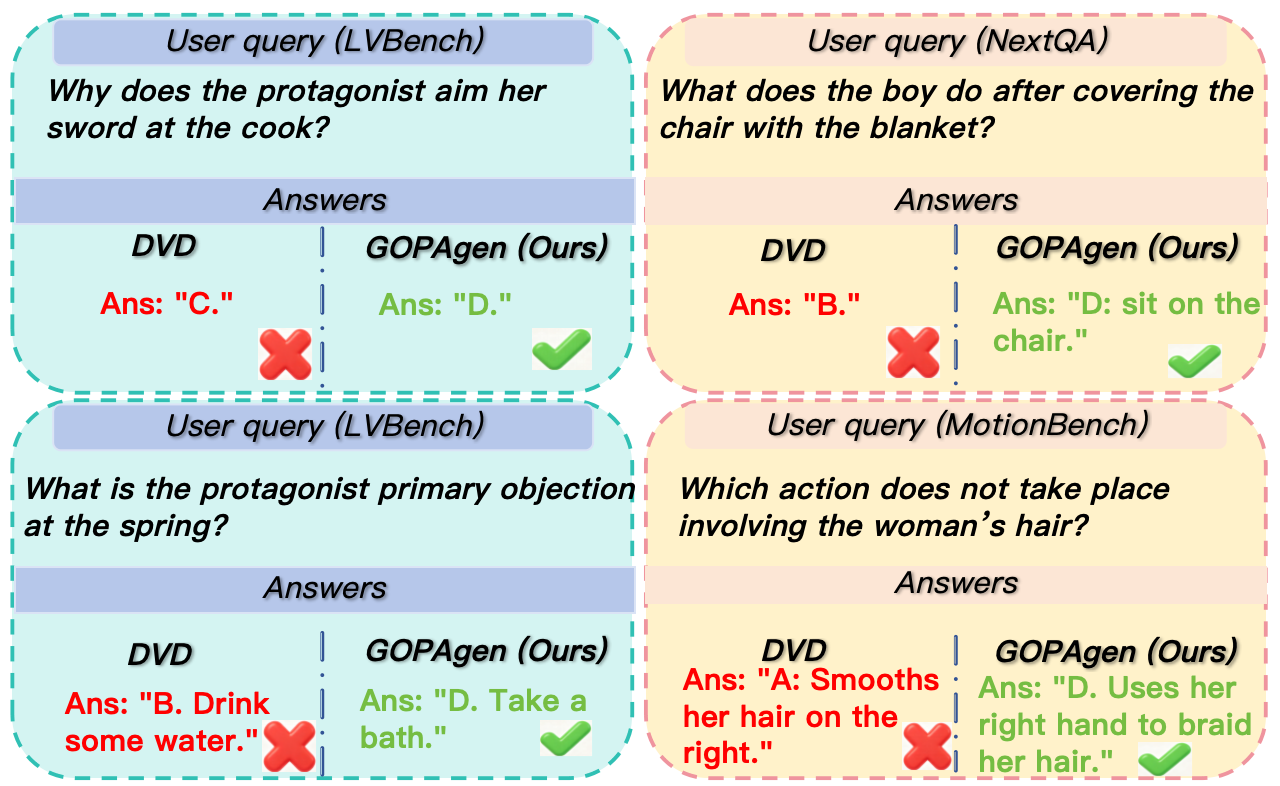}
    \caption{Additional answer comparisons with Deep Video Discovery (DVD). LVBench examples test character motivation and intended activity; NExT-QA tests action order; MotionBench tests identification of an action that does not occur. Model responses and their correctness markers are shown for each question.}
    \label{fig:app4}
\end{figure}

\begin{figure}[t]
    \centering
    \includegraphics[width=\linewidth]{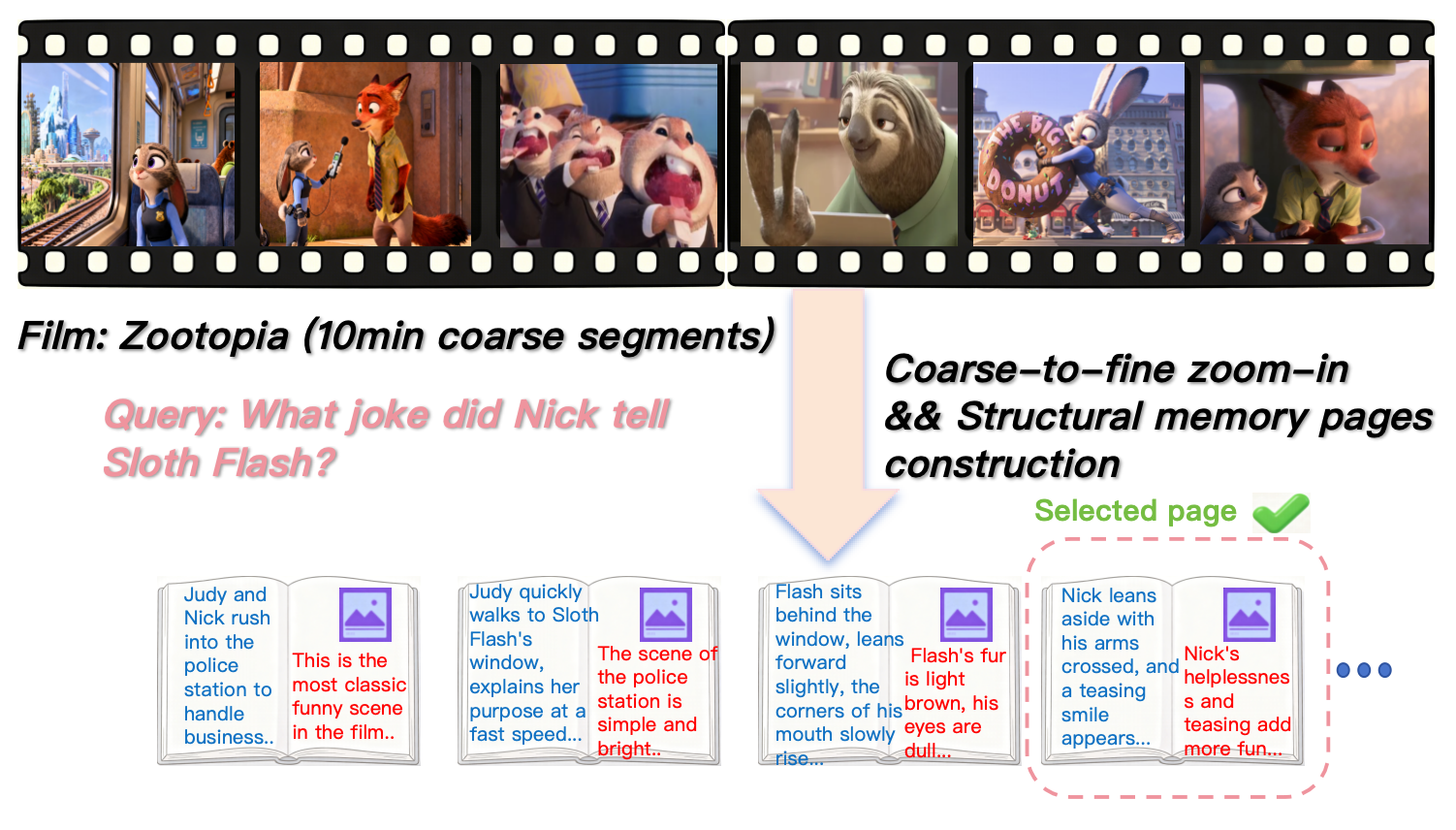}
    \caption{Qualitative memory-selection illustration using a scene from \emph{Zootopia}. The question asks what joke Nick tells Flash. Film frames provide visual context, while the lower row displays textual memory pages. A dashed outline and check mark identify the selected page; the figure illustrates page selection rather than a benchmark accuracy measurement.}
    \label{fig:app5}
\end{figure}

\begin{figure}[t]
    \centering
    \includegraphics[width=\linewidth]{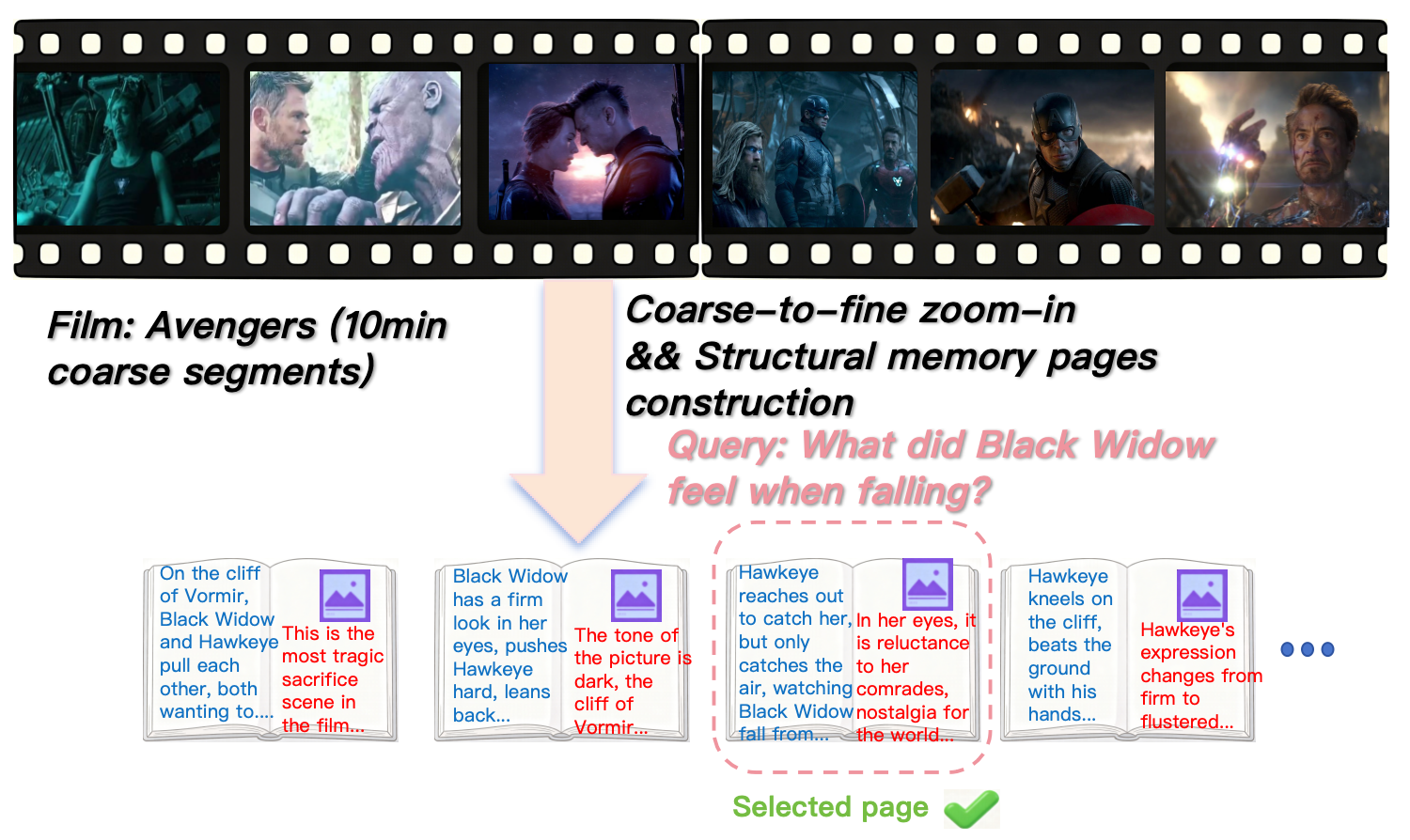}
    \caption{Qualitative memory-selection illustration using an \emph{Avengers} scene. The question concerns Black Widow's feelings while falling. The lower row shows candidate textual descriptions, with the selected memory page highlighted. The example illustrates the retrieved representation and does not establish that the character's internal state is directly observable.}
    \label{fig:app6}
\end{figure}

\begin{figure}[t]
    \centering
    \includegraphics[width=0.5\linewidth]{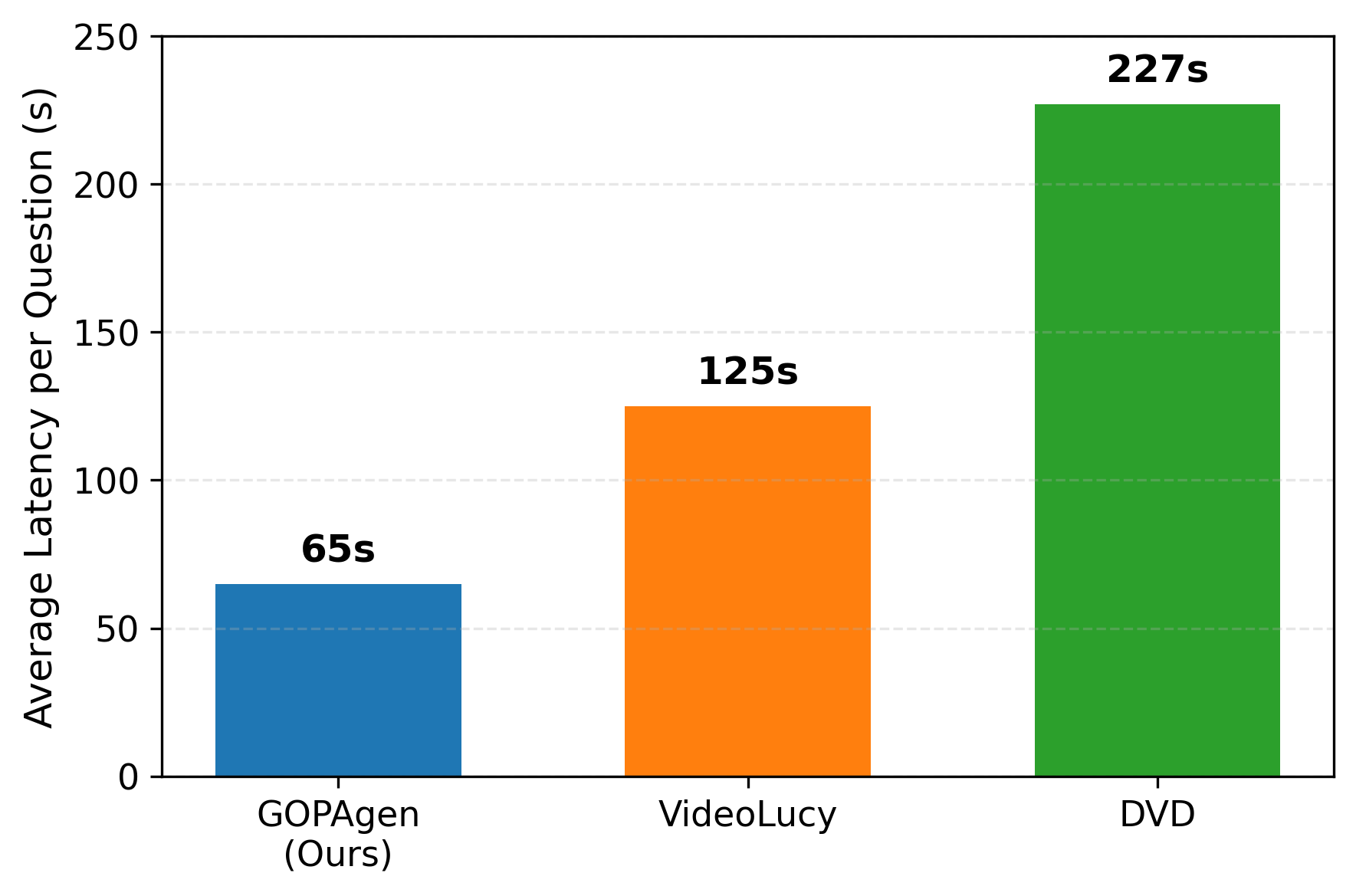}
    \caption{Average query-processing latency on 200 LVBench validation videos spanning 10--60 minutes, with one question per video and three runs per method. GOPAgen, VideoLucy, and Deep Video Discovery take 65, 125, and 227 seconds, respectively. Runs use one NVIDIA A100 80-GB GPU, 64 CPU cores, one concurrent query, and a cold application cache. Latency includes inference, retrieval, and API waiting but excludes one-time preprocessing. Dispersion and API details are reported in~\cref{app:preliminary_time_efficiency}.}
    \label{fig:latency}
\end{figure}

\begin{figure*}[t]
    \centering
    \includegraphics[width=0.8\linewidth]{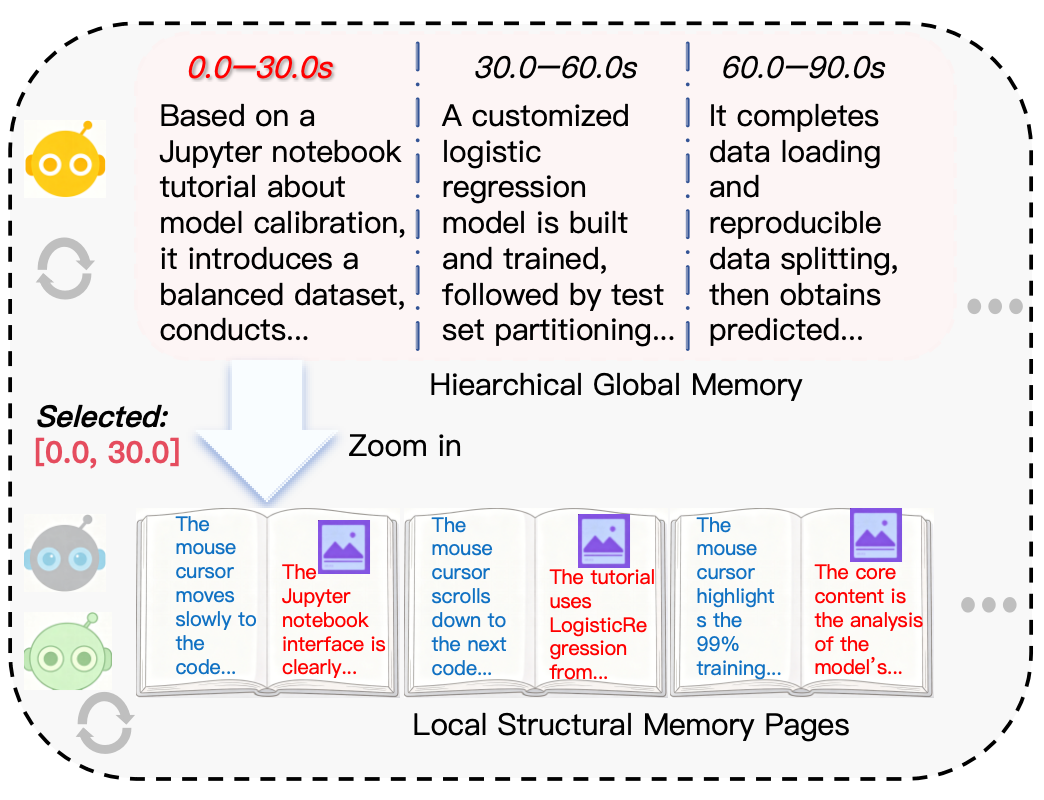}
    \vspace{-0.1in}
    \caption{Global-to-local memory construction for a tutorial video. The upper block summarizes successive displayed time windows covering 0--90 seconds. The 0--30-second region is selected for closer inspection, and the lower block shows local pages describing screen activity and visual content. Arrows connect coarse textual memory to the more detailed page representation; this is a memory-construction example, not a model-answer comparison.}
    \label{fig:444}
    \vspace{-0.2in}
\end{figure*}

Figures~\ref{fig:app1}--\ref{fig:app4} compare model answers, and~\cref{fig:app5,fig:app6} illustrate memory selection for film scenes. Figure~\ref{fig:444} shows global-to-local memory construction for a tutorial video. These examples complement the quantitative results by distinguishing answer comparisons from memory visualizations. The latency comparison in~\cref{fig:latency} follows the protocol in~\cref{app:preliminary_time_efficiency}.


\section{Broader Impacts}
\label{app:broader_impacts}
GOPAgen studies selective long-video processing through codec-aware evidence and structured memory. Its resource measurements characterize the reported hardware, workload, and caching conditions; they do not establish a hardware-independent reduction in energy consumption.

Long-video analysis can raise privacy concerns, particularly in surveillance and monitoring. Responsible deployment requires appropriate consent, access controls, and attention to applicable privacy requirements. Biases in datasets and external models may also produce inaccurate or unfair outputs. Retrieved descriptions and model-generated answers therefore require scrutiny in consequential applications.

\section{Limitations}
\label{app:limitations}
GOPAgen is a modular system rather than a unified end-to-end learned model. Accuracy and cost depend on the global and caption agents, motion agent, ranker, embedding model, database, and external reasoning service. Comparisons across systems remain sensitive to model capacity and inference budgets.

Codec motion vectors can be noisy around scene cuts, occlusions, small objects, and camera motion. Fixed-depth temporal selection is heuristic, and translating visual evidence into text can discard information. The component, retrieval, sensitivity, and failure analyses characterize these limitations under the evaluated settings.

GOPAgen remains below the strongest listed baseline on LVBench and Video-MME Long. Its measured efficiency is specific to the reported workloads and does not imply the same advantage under different preprocessing, cache, or multi-query conditions.

\end{document}